\documentclass[10pt,journal,compsoc]{IEEEtran}

\ifCLASSOPTIONcompsoc
  \usepackage[nocompress]{cite}
\else
  \usepackage{cite}
\fi
\ifCLASSINFOpdf
\else
\fi

\usepackage{amsmath}
\usepackage{amssymb}
\usepackage{algorithmic}
\usepackage{algorithm}
\usepackage{enumerate}
\usepackage{graphicx}
\usepackage{color}
\usepackage{tikz}

\usepackage{colortbl}

\ifCLASSOPTIONcompsoc
  \usepackage[caption=false,font=footnotesize,labelfont=sf,textfont=sf]{subfig}
\else
  \usepackage[caption=false,font=footnotesize]{subfig}
\fi

\usepackage{booktabs}
\usepackage{dblfloatfix}

\usepackage{url}
\usepackage{bm}
\usepackage{hyperref}
\usepackage{multirow,tabularx}
\newcolumntype{C}{>{\centering\arraybackslash}X}

\usepackage{ulem}

\usepackage{amsthm}

\theoremstyle{plain}

\theoremstyle{definition}

\usepackage{xparse}

\newenvironment{colored*}{\begingroup\color{black}}{\endgroup} 
\newcommand{\best}[1]{\cellcolor[HTML]{ECC0BD}#1}
\newcommand{\second}[1]{\cellcolor[HTML]{F4D9D7}#1}
\newcommand{\third}[1]{\cellcolor[HTML]{FBF2F1}#1}
\definecolor{bestcolor}{HTML}{ECC0BD}
\definecolor{secondcolor}{HTML}{F4D9D7}
\definecolor{thirdcolor}{HTML}{FBF2F1}
\newcommand{\colorswatch}[1]{\tikz[baseline=-0.35em]{\fill[#1] (0,0) circle (0.45em);}}

\begin{document}
%
\title{\huge ReX-Shot: Single-Image Rephotography via Geometry- and Camera-Grounded Generation}

%
%
%
%

\author{Ruiqi Zhang$^\dagger$, Hao Zhu$^\dagger$, Wenhao Zhang, Qi Zhang, Junqi Shi, Ming Lu, Xun Cao, Zhan Ma
\IEEEcompsocitemizethanks{\IEEEcompsocthanksitem The first two authors contributed equally. 
\IEEEcompsocthanksitem R. Zhang, H. Zhu, W. Zhang, J. Shi, M. Lu, X. Cao, Z. Ma are with the School of Electronic Science and Engineering, Nanjing University, Nanjing, 210023, China.
\IEEEcompsocthanksitem Q. Zhang is with vivo Mobile Communication Co., Ltd., Xi'an, 710055, China.
}
\thanks{\par\vspace*{0.2\baselineskip}\noindent%
Manuscript received April 19, 2005; revised August 26, 2015.}}

%
%

\markboth{Journal of \LaTeX\ Class Files,~Vol.~14, No.~8, August~2015}%
{Shell \MakeLowercase{\textit{et al.}}: Bare Advanced Demo of IEEEtran.cls for IEEE Computer Society Journals}

\IEEEtitleabstractindextext{%
\begin{abstract}
Single-image rephotography aims to synthesize new shots of the same scene from a single reference image with specified viewpoints, focal lengths, and parameterized photographic effects, all of which are intrinsically coupled in the imaging process. However, existing methods typically treat them as separate subproblems and struggle under joint control, where novel view synthesis may distort geometry under focal-length changes, while super resolution and instruction-guided editing remain confined to 2D and cannot extend detail restoration or appearance control to novel viewpoints. We find that these limitations arise from two fundamental issues, namely the ill-posed nature of single-image 3D reconstruction and the sampling limitation of continuous focal-length enlargement. The former makes geometric inaccuracies unavoidable during pixel-space explicit projection, while the latter requires recovering details beyond the reference image's sampling density. To mitigate projection bias from imperfect geometry, we leverage implicitly transformed foundation-model features to provide robust target-view guidance. We then formulate focal-length enlargement as a geometry-guided super-resolution problem, using generative detail priors to recover details missing from sparse 3D resampling. Building on this 3D-aware generative backbone, we lift photographic-effect control from 2D filtering to 3D-aware appearance editing, preserving content consistency across viewpoints and focal lengths. Together, these components form ReX-Shot, a geometry- and camera-grounded generative framework for single-image rephotography. To the best of our knowledge, ReX-Shot is the first unified framework to jointly control viewpoint, focal length, and parameterized photographic effects from a single reference image. Experiments demonstrate that ReX-Shot outperforms representative baselines across viewpoint control, focal-length control, and photographic-effect control, while enabling interactive rephotography at near-real-time speeds. \textit{Project page:} \url{https://ruiqi-nju.github.io/ReX-Shot/}
\end{abstract}

}

\maketitle

\IEEEdisplaynontitleabstractindextext

%
\IEEEpeerreviewmaketitle

\ifCLASSOPTIONcompsoc
\IEEEraisesectionheading{\section{Introduction}\label{sec:introduction}}
\else
\section{Introduction}
\label{sec:introduction}
\fi

\begin{figure*}[!t]
    \centering
    \includegraphics[width=\textwidth]{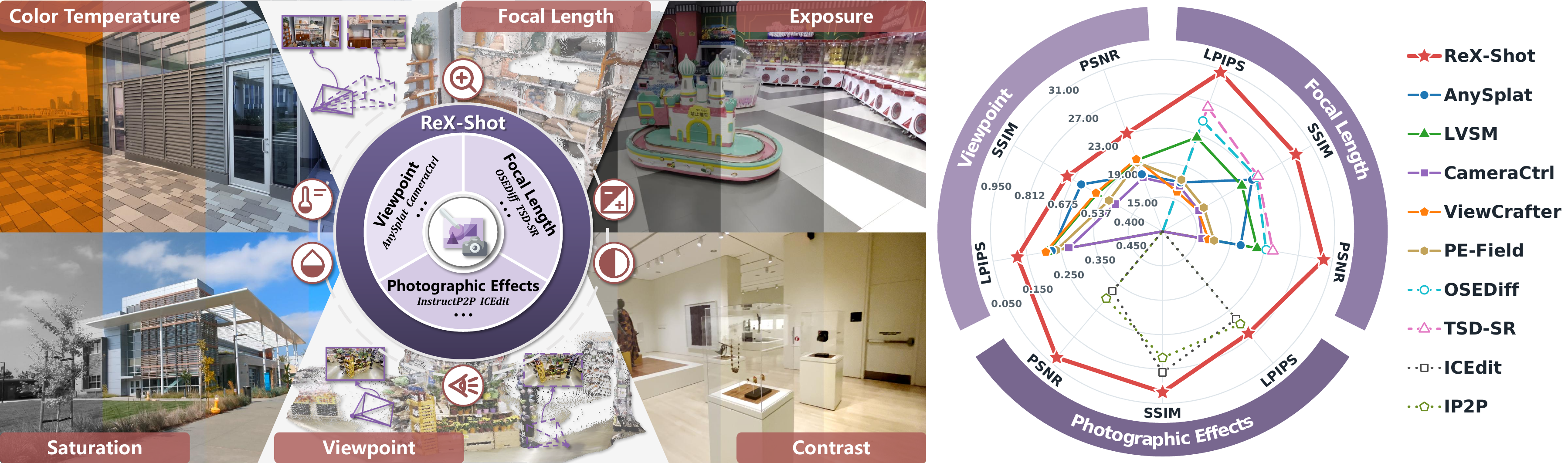}
    \caption{\textbf{Overview of ReX-Shot for single-image rephotography.} \textbf{Left: Multidimensional rephotography capabilities.} ReX-Shot virtually recaptures the reference scene as target shots with controllable camera geometry and photographic effects, including viewpoint, focal length, color temperature, exposure, saturation, and contrast. \textbf{Right: Quantitative performance comparison.} ReX-Shot supports viewpoint, focal-length, and photographic-effect control while outperforming representative baselines from different method categories. All results in the radar chart are evaluated on the DL3DV-140~\cite{Ling_2024_CVPR} test set. For each metric, we globally normalize the scores across the three tasks, and the numerical labels on the concentric grid rings shown in the Viewpoint section also apply to the same metric in the Focal Length and Photographic Effects sections. The thick solid, thin solid, dashed, and dotted lines denote ReX-Shot, Novel View Synthesis, Super Resolution, and Instruction-Guided Editing, respectively.}
    \label{fig:first}
\end{figure*}

\IEEEPARstart{P}{hotography} is fundamentally a constrained image-formation process that maps the high-dimensional physical world onto a static two-dimensional sensor plane. Within this process, camera geometry (viewpoint and focal length) determines projection and composition, and photographic effects determine image appearance. Once the shutter button is pressed, these settings are fixed in the captured image. Therefore, conventional post-processing cannot ensure scene consistency as if it were recaptured under new camera settings. To address this limitation, we define single-image rephotography as synthesizing new shots of the same scene from a single reference image with specified viewpoints, focal lengths, and parameterized photographic effects, such as color temperature, exposure, saturation, and contrast. Unlike generic image editing, rephotography requires preserving scene content and spatial relationships while faithfully responding to the target camera geometry and photographic-effect parameters.

Existing methods typically treat viewpoint control, focal-length control, and photographic-effect control as separate subproblems. However, these subproblems are tightly coupled in rephotography, as spatial composition and image appearance must be controlled jointly to produce a coherent final capture. For viewpoint control, novel view synthesis and camera-aware generation methods~\cite{sargent2024zeronvs,gao2024cat3d,chung2025luciddreamer,wang2024motionctrl,jiang2025anysplat,jin2025lvsm,yu2025viewcrafter,bai2026pefield,he2024cameractrl} synthesize target views from sparse or single inputs. These methods have demonstrated strong novel-view generation capabilities and effective camera motion control. When directly applied to focal-length control, however, they may produce incorrect fields of view, object deformation, blurred details, or unfaithful over-generation artifacts. For focal-length control, generative super-resolution and restoration methods~\cite{wang2024stablesr,wu2024seesr,lin2024diffbir,yu2024supir,wu2024osediff,dong2025tsdsr} are effective at enhancing enlarged images and recovering fine details, but they do not ensure geometric consistency during continuous focal-length changes. For photographic effects, instruction-guided image editing methods~\cite{Brooks_2023_CVPR,zhang2023magicbrush,Sheynin_2024_CVPR,yu2024anyedit,fu2024mgie,zhang2025icedit} can modify image appearance, and recent generative photography methods further explore numerical or parameterized control~\cite{fang2024camera,Yuan_2025_CVPR,qin2025camedit,yang2025cameramaster}, but they remain largely confined to the 2D image plane and cannot be integrated with camera-geometry control. Consequently, despite substantial progress along each individual dimension, jointly controlling viewpoint, focal length, and photographic effects within a unified framework for single-image rephotography remains unexplored.

These limitations arise from two fundamental issues. First, 3D reconstruction from a single image is inherently ill-posed~\cite{Yin_2023_ICCV}, as the same 2D observation can correspond to multiple plausible 3D geometries. Consequently, geometry estimation errors are unavoidable and can lead to inaccurate projections of object shapes and spatial relationships onto the target image plane. Second, focal-length enlargement is sampling limited, as continuous focal-length changes introduce different degrees of sparsity in the target image, requiring details increasingly beyond the original sampling density of the reference image. Therefore, focal-length control must recover these missing details while preserving geometric consistency across this range. These issues further hinder the establishment of a stable 3D-aware backbone for integrating photographic effects with camera geometry, even though such effects are key elements of real photography and often depend heavily on photographic experience.

To address the aforementioned issues, we introduce two key designs. First, we propose dual-space geometry and camera grounding. We observe that, although recent state-of-the-art visual geometry models~\cite{wang2024dust3r,wang2025vggt,wang2025pi3} provide strong scene reconstruction performance, geometric inaccuracies remain unavoidable, and even small structural errors can be amplified into noticeable local offsets during pixel-space explicit projection. To mitigate this pixel-space projection bias, we introduce feature-space implicit projection that learns an implicit transformation of VGGT and DINOv2~\cite{oquab2023dinov2} features from the reference view to the target view. Leveraging robust foundation-model features, this projection provides context-aggregated semantic and geometric cues beyond explicit point-wise correspondences, leading to more stable target-view guidance under imperfect geometric projection.

Second, we observe that increasing the focal length reduces the spatial sampling density, making detail recovery naturally related to super resolution. Therefore, to address the sampling limitation, we leverage the detail prior of a super-resolution model~\cite{wang2024stablesr,lin2024diffbir,wu2024seesr,dong2025tsdsr}, which has been used in 2D image enhancement but is rarely explored for alleviating sparsity caused by 3D scene resampling. We fine-tune a one-step generative super-resolution model with dual-space conditions, allowing missing details to be recovered with geometric guidance while preserving inference efficiency.

These two designs establish a 3D-aware generative backbone that supports content-consistent synthesis across viewpoints and focal lengths. Building on this backbone, we further incorporate photographic-effect control. Specifically, we continuously parameterize each photographic effect and encode it with a dedicated encoder, whose control signal is then injected into the backbone. In this way, photographic-effect control is lifted from conventional 2D filtering to 3D-aware appearance editing that preserves content consistency across viewpoints and focal lengths.

Together, these components form ReX-Shot, the first unified framework for single-image rephotography, as illustrated in Fig.~\ref{fig:first}. Our contributions are summarized as follows:
\begin{itemize}
    \item We define single-image rephotography as a virtual reshooting process, which recaptures the reference scene with specified viewpoints, focal lengths, and parameterized photographic effects.
    \item We identify two fundamental issues behind single-image rephotography, namely the ill-posed nature of single-image 3D reconstruction and the sampling limitation under continuous focal-length enlargement.
    \item We propose two key designs to address these issues, \textit{i.e.}, feature-space implicit projection for mitigating projection bias caused by imperfect geometry, and a generative super-resolution prior for geometry-guided restoration under focal-length changes.
    \item We substantiate that ReX-Shot outperforms representative baselines across viewpoint control, focal-length control, and photographic-effect control, while enabling near-real-time interactive rephotography.
\end{itemize}

\section{Related Work}
\subsection{Camera-Controlled View Synthesis}
Novel view synthesis has progressed from per-scene optimized 3D representations to generalizable reconstruction and view generation models. Classical neural rendering methods, such as NeRF~\cite{mildenhall2020nerf} and 3D Gaussian Splatting~\cite{kerbl20233d}, optimize a scene-specific radiance field or Gaussian representation from posed images and render novel views from the learned scene. While effective, such per-scene optimization typically requires multiple observations and is not directly suited to single-image rephotography. Beyond per-scene optimization, recent methods have explored predicting scene representations or target views from sparse or single inputs. Representative feed-forward Gaussian reconstruction methods, including pixelSplat~\cite{charatan2024pixelsplat}, MVSplat~\cite{chen2024mvsplat}, and AnySplat~\cite{jiang2025anysplat}, predict explicit Gaussian scene representations from image pairs, sparse multi-view inputs, or unconstrained views, retaining a renderable 3D proxy while improving generalization. Transformer-based novel-view-synthesis methods such as LVSM~\cite{jin2025lvsm} synthesize target views from sparse posed images, while RayZer~\cite{jiang2025rayzer} follows the same line by jointly estimating cameras and scene representations from unposed and uncalibrated inputs.

Recent generative models also expose camera geometry as a condition for view generation. MotionCtrl~\cite{wang2024motionctrl} controls video generation with camera pose sequences, while CameraCtrl~\cite{he2024cameractrl} and CAT3D~\cite{gao2024cat3d} condition diffusion generation with dense camera ray maps. Stable Virtual Camera~\cite{zhou2025stablevirtualcamera} generalizes this setting to any number of input views and target cameras. Another line of geometry-guided generative methods uses 3D cues to steer diffusion-based view generation. Methods with explicit geometric conditions, including GenWarp~\cite{seo2024genwarp}, MultiDiff~\cite{muller2024multidiff}, and ViewCrafter~\cite{yu2025viewcrafter}, warp or project the reference image with estimated depth or point-based geometry before using diffusion priors to complete missing content. Geometry-aware generators such as GeNVS~\cite{chan2023genvs} and PE-Field~\cite{bai2026pefield} instead inject 3D structure into the generative model through latent 3D feature volumes or 3D positional encodings.

These methods substantially advance controllable view generation, but they mainly target viewpoint or camera-trajectory control and do not explicitly handle focal-length changes. ReX-Shot instead represents viewpoint and focal length jointly as target camera geometry, grounding generation through pixel-space explicit projection and feature-space implicit projection.

\subsection{Image Editing and Generative Photography}
Generative image editing provides a flexible route for controllable appearance modification by translating natural-language instructions into image changes. Early instruction-guided methods mainly obtain edit supervision from paired editing data, with InstructPix2Pix~\cite{Brooks_2023_CVPR} learning from synthetic instruction-image pairs and MagicBrush~\cite{zhang2023magicbrush} improving this paradigm with manually annotated real-image editing triplets. Subsequent work strengthens the data-driven formulation along two complementary directions. Emu Edit~\cite{Sheynin_2024_CVPR} broadens the supervision signal with recognition-and-generation tasks and a richer taxonomy of editing operations, and AnyEdit~\cite{yu2024anyedit} further scales high-quality multimodal instruction-editing pairs to cover diverse edit types and improve cross-task generalization. In parallel, methods that enhance instruction understanding use richer language or multimodal context to handle more complex user intents. MGIE~\cite{fu2024mgie} leverages multimodal large language models to produce expressive editing guidance, while ICEdit~\cite{zhang2025icedit} formulates image editing as in-context generation in a large-scale diffusion transformer.

Generative photography further studies how to expose photographic controls to generative models. Camera Settings as Tokens~\cite{fang2024camera} encodes numerical photographic-effect parameters into latent diffusion models, enabling both text-to-image generation and image editing under specified photographic settings. Generative Photography~\cite{Yuan_2025_CVPR} focuses on text-to-image synthesis and improves scene-consistent photographic control through camera-condition embeddings. In the image-to-image setting, CamEdit~\cite{qin2025camedit} introduces continuous prompting with photographic-effect parameters and parameter-aware modulation for fine-grained appearance editing. CameraMaster~\cite{yang2025cameramaster} further explores unified photographic retouching by incorporating both semantic camera directives and numerical parameter conditions for compositional control over multiple photographic effects.

Instruction-guided editing supports broad semantic changes but lacks precise parameterized control over photographic effects. Generative photography moves closer by conditioning on numerical photographic-effect parameters, yet most methods still rely on the disentanglement of scene content and photographic-effect parameters within textual instructions and remain largely confined to the 2D image plane. In contrast, ReX-Shot anchors scene content with deterministic geometry and composes parameterized photographic-effect control with viewpoint and focal-length changes.

\begin{figure*}[!t]
    \centering
    \includegraphics[width=\textwidth]{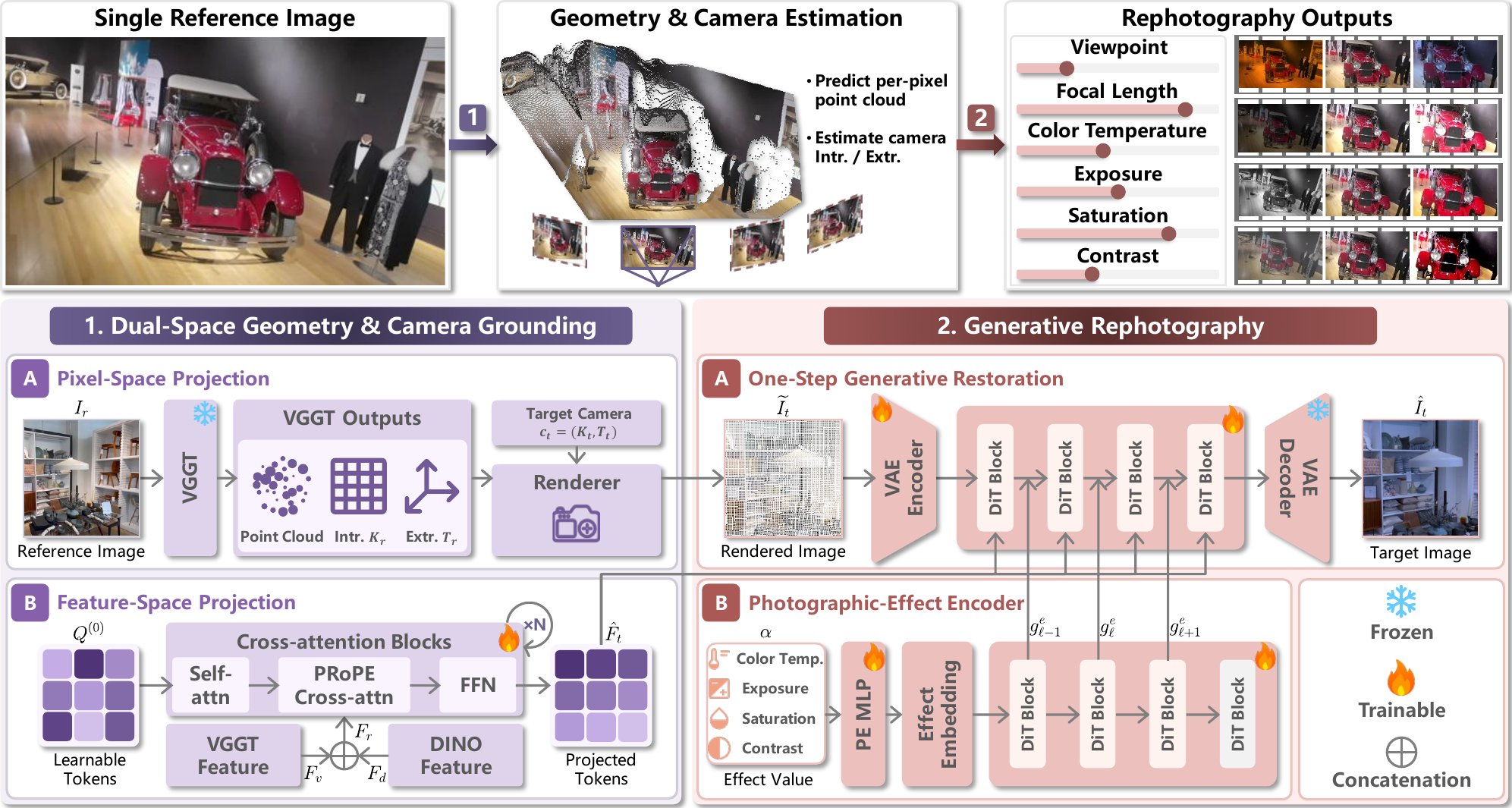}
    \caption{\textbf{Pipeline of ReX-Shot.}
    Given a single reference image, ReX-Shot first estimates a per-pixel point cloud and camera parameters, and then synthesizes rephotographed images with controllable viewpoint, focal length, and photographic effects.
    The pipeline comprises two stages.
    \textbf{1. Dual-space geometry and camera grounding} combines pixel-space projection, which renders the estimated point cloud under the target camera, with feature-space projection, which implicitly transforms VGGT~\cite{wang2025vggt} and DINOv2~\cite{oquab2023dinov2} features into camera-aware tokens through PRoPE-based cross-attention~\cite{li2025prope}.
    \textbf{2. Generative rephotography} leverages the detail prior of a one-step generative super-resolution model to recover details missing from sparse 3D resampling under the guidance of dual-space conditions, while a parameterized photographic-effect encoder injects continuous effect controls into the DiT blocks. Snowflakes and flames denote frozen and trainable modules, respectively, and $\oplus$ denotes concatenation.}
    \label{fig:pipeline}
\end{figure*}

\subsection{Diffusion Priors for Restoration}
Diffusion priors have become a strong backbone for real-world image super resolution and restoration. Representative methods leverage pretrained text-to-image diffusion models as image priors for detail recovery. Among them, StableSR~\cite{wang2024stablesr} balances fidelity and perceptual quality with a time-aware encoder, DiffBIR~\cite{lin2024diffbir} combines an initial reconstruction network with Stable Diffusion~\cite{Rombach_2022_CVPR} detail synthesis, and SeeSR~\cite{wu2024seesr} improves semantic fidelity with degradation-aware prompts. More recent works reduce inference cost through diffusion distillation and one-step formulations. SinSR~\cite{wang2023sinsr} uses consistency-preserving distillation to perform super resolution in a single step, AddSR~\cite{xie2024addsr} introduces adversarial diffusion distillation for blind super resolution, and OSEDiff~\cite{wu2024osediff} starts diffusion directly from the low-quality input to avoid stochastic uncertainty. TSD-SR~\cite{dong2025tsdsr} further introduces target score distillation for stronger detail recovery and faster inference.

Generative priors have also been used to compensate for incomplete 3D evidence in sparse-view reconstruction and novel view synthesis. ReconFusion~\cite{wu2023reconfusion} regularizes few-view NeRF reconstruction with diffusion-based novel-view priors, while ViewCrafter~\cite{yu2025viewcrafter} combines coarse point-based geometry with a video diffusion prior to synthesize high-fidelity novel views and progressively extend the covered view range. ReconX~\cite{liu2024reconx} constructs a global point cloud from sparse input views, uses it as a 3D structure condition for video diffusion, and then optimizes a 3D Gaussian Splatting representation from the generated sequence. For 3D Gaussian Splatting restoration, DiFiX3D+~\cite{wu2025difix3d} and GSFixer~\cite{yin2025gsfixer} use diffusion priors to repair artifacts caused by underconstrained or degraded 3D representations.

ReX-Shot draws insights from these two categories of methods for single-image rephotography. Generative priors are useful for both recovering fine details required by focal-length enlargement and synthesizing missing novel-view content when explicit 3D evidence is incomplete. ReX-Shot therefore adapts a generative super-resolution model to perform target-camera detail restoration and content completion.

\section{Method}
\label{sec:method}

Fig.~\ref{fig:pipeline} provides an overview of the ReX-Shot pipeline. In the following, we first formulate single-image rephotography in Section~\ref{sec:problem_formulation}, and then introduce dual-space geometry and camera grounding in Section~\ref{sec:dual_space_grounding}, comprising pixel-space explicit projection in Section~\ref{sec:pixel_reprojection} and feature-space implicit projection in Section~\ref{sec:feature_grounding}. We subsequently present generative rephotography with a super-resolution (SR) prior in Section~\ref{sec:diffusion_completion}, and parameterized photographic-effect control in Section~\ref{sec:effect_control}.

\subsection{Single-Image Rephotography}
\label{sec:problem_formulation}
Given a reference image $I_r \in [0,1]^{H \times W \times 3}$, ReX-Shot synthesizes a target image $\hat I_t$ under a new shot specification. We denote the reference and target cameras by $c_r=(K_r,T_r)$ and $c_t=(K_t,T_t)$, respectively. Each camera is parameterized by an intrinsic matrix $K$ and an extrinsic matrix $T=[R \mid t]$, where $R$ and $t$ denote rotation and translation, respectively. The target camera $c_t$ specifies the desired shot by coupling focal-length control through $K_t$ with viewpoint control through $T_t$. When photographic-effect control is requested, we further use an effect condition $e_\phi=\alpha$, where $\phi$ denotes the photographic-effect type and $\alpha\in[0,1]$ its normalized control value. The task is therefore written as
\begin{equation}
    \hat I_t = \mathcal{F}_{\Theta}(I_r,c_t,e_\phi),
    \quad e_\phi=\varnothing \ \text{when no effect is applied}.
    \label{eq:rexshot_task}
\end{equation}
This formulation defines rephotography as target-shot formation governed by camera geometry and parameterized photographic effects. The model must preserve scene content and spatial relationships under the target viewpoint, recover details affected by focal-length-induced resampling, and realize the requested photographic effect when provided.

ReX-Shot realizes Eq.~\eqref{eq:rexshot_task} through two stages, as illustrated in Fig.~\ref{fig:pipeline}. The first stage establishes dual-space geometry and camera grounding. In pixel space, we use VGGT~\cite{wang2025vggt} to estimate a per-pixel point cloud from the reference image and reproject it under the target camera to produce a coarse geometry anchor. In feature space, we design a camera-aware feature projector that implicitly transforms VGGT and DINOv2~\cite{oquab2023dinov2} features according to the relative projective relation between the reference and target cameras. The second stage feeds these complementary conditions to a one-step diffusion model initialized from TSD-SR~\cite{dong2025tsdsr}, which completes missing regions and recovers fine details. Within the second stage, we introduce a dedicated encoder that injects the effect condition into the generative backbone to control the photographic effect.

\subsection{Dual-Space Geometry and Camera Grounding}
\label{sec:dual_space_grounding}

\subsubsection{Pixel-Space Explicit Projection}
\label{sec:pixel_reprojection}
The pixel-space branch explicitly projects the reconstructed scene into the target camera. Given the reference image $I_r$, VGGT~\cite{wang2025vggt} predicts its depth map $D_r$ and camera parameters $(K_r,T_r)$. For each reference pixel $u=(x,y,1)^\top$, we back-project it to a 3D point in the world coordinate system and project the point onto the target image plane:
\begin{equation}
    \begin{aligned}
        X(u) &= R_r^\top\left(D_r(u)K_r^{-1}u-t_r\right), \\
        u_t(u) &= \pi\!\left(K_t\left(R_tX(u)+t_t\right)\right),
    \end{aligned}
    \label{eq:point_cloud_projection}
\end{equation}
where $\pi([p_x,p_y,p_z]^\top)=(p_x/p_z,p_y/p_z)^\top$ denotes perspective division. We rasterize the projected points with z-buffering to obtain a coarse target rendering $\widetilde I_t$. Because the point cloud contains only samples visible in the reference image, $\widetilde I_t$ can become incomplete due to viewpoint-induced disocclusion and sparse due to focal-length enlargement. More importantly, unavoidable inaccuracies in single-image geometry estimation can introduce local projection offsets in $\widetilde I_t$. Consequently, it serves only as a coarse geometry anchor for subsequent generative completion and detail recovery.

\subsubsection{Feature-Space Implicit Projection}
\label{sec:feature_grounding}
Pixel-space projection provides an explicit geometry anchor, but its point-wise correspondences are sensitive to local offsets caused by unavoidable geometry inaccuracies. We therefore complement it with an implicit feature-space projection branch. Specifically, given $I_r\in[0,1]^{H\times W\times 3}$, we extract the VGGT~\cite{wang2025vggt} and DINOv2~\cite{oquab2023dinov2} feature maps as
\begin{equation}
    \begin{aligned}
        F_v &= \mathcal{E}_{\mathrm{VGGT}}(I_r)
        \in\mathbb{R}^{\frac{H}{14}\times\frac{W}{14}\times d_v}, \\
        F_d &= \mathcal{E}_{\mathrm{DINO}}(I_r)
        \in\mathbb{R}^{\frac{H}{14}\times\frac{W}{14}\times d_d},
    \end{aligned}
    \label{eq:feature_extraction}
\end{equation}
where $d_v$ and $d_d$ denote the respective channel dimensions. Following the patch-based tokenization of VGGT and DINOv2, each spatial location in these feature maps corresponds to a $14\times14$ image patch rather than an individual pixel. Moreover, interactions among tokens at different spatial locations allow each feature to aggregate information from a broader image region. These properties make the resulting representation more robust to local projection offsets. To combine the geometry-aware scene structure encoded by VGGT with the robust visual semantics captured by DINOv2, we concatenate the two feature maps channel-wise to obtain $F_r=[F_v;F_d]\in\mathbb{R}^{\frac{H}{14}\times\frac{W}{14}\times(d_v+d_d)}$.

Changes in viewpoint and focal length alter scene visibility and the field of view, making some reference-view features irrelevant to the target view. We therefore design a camera-aware feature projector to transform $F_r$ into target-view tokens while filtering irrelevant reference content. Specifically, the projector applies a stack of transformer blocks with PRoPE cross-attention~\cite{li2025prope}, which encodes the reference and target camera parameters as relative positional encodings:
\begin{equation}
\begin{aligned}
    Q^{(\ell)}
    &= \mathcal{B}_{\psi}^{(\ell)}
    \left(Q^{(\ell-1)};F_r,\mathcal{P}\right),
    \quad \ell=1,\ldots,L_p, \\
    \mathcal{B}_{\psi}^{(\ell)}
    &= \mathrm{FFN}^{(\ell)}
    \circ \mathrm{CA}_{\mathrm{PRoPE}}^{(\ell)}
    \circ \mathrm{SA}^{(\ell)},
    \quad \hat F_t=Q^{(L_p)},
\end{aligned}
    \label{eq:feature_projector}
\end{equation}
where $Q^{(0)}$ denotes a set of learnable query tokens that are randomly initialized before training, $\mathcal{P}=(c_r,c_t)$ denotes the reference--target camera pair, $\mathcal{B}_{\psi}^{(\ell)}$ is the $\ell$-th transformer block, and $L_p$ is the number of projector blocks. The operators $\mathrm{SA}$, $\mathrm{CA}_{\mathrm{PRoPE}}$, and $\mathrm{FFN}$ denote self-attention, PRoPE cross-attention, and the feed-forward network, respectively, and $\hat F_t$ denotes the projected target-view tokens. This transformation is performed entirely in feature space and does not require explicit scene geometry. It is therefore less sensitive to local geometry errors and provides more stable target-view guidance.

However, if we jointly train the feature projector with the generative backbone using only image-level supervision, it may fail to learn the intended camera transformation, forcing the generative backbone to locate target-camera-relevant cues directly within the reference-view features. We therefore pretrain the projector against paired target-view features $F_t$, which are extracted from the paired target image $I_t$ using the same frozen VGGT and DINOv2 encoders as $F_r$. This direct feature-level supervision encourages $\hat F_t$ to align with the target camera before it is used to guide generation. For $N$ patch tokens, we minimize the cosine similarity loss
\begin{equation}
    \mathcal{L}_{\mathrm{align}}
    =
    \frac{1}{N}\sum_{i=1}^{N}
    \left(
    1-
    \frac{\left\langle \hat F_t^{(i)},F_t^{(i)}\right\rangle}
    {\left\|\hat F_t^{(i)}\right\|_2\left\|F_t^{(i)}\right\|_2}
    \right),
    \label{eq:feature_alignment_loss}
\end{equation}
while keeping the VGGT and DINOv2 encoders frozen. This objective trains the projector to map reference-view features to target-camera-aligned tokens.

\subsection{Generative Rephotography with an SR Prior}
\label{sec:diffusion_completion}
Focal-length enlargement creates a detail-recovery problem closely related to super-resolution (SR). Let $\rho_r$ and $\rho_t$ denote the densities of available reference samples before and after reprojection, respectively. Under focal-length enlargement, the spatial coordinates are scaled by $s_f=f_t/f_r$, and the corresponding sample density becomes
\begin{equation}
    \rho_t
    =
    \frac{\rho_r}{s_f^2},
    \qquad
    s_f=\frac{f_t}{f_r}>1.
    \label{eq:focal_density_relation}
\end{equation}
An SR model with scale factor $s$ reconstructs $s^2$ output samples per input sample, corresponding to an output sample density
\begin{equation}
    \rho_{\mathrm{HR}}
    =
    s^2\rho_{\mathrm{LR}},
    \label{eq:sr_density_relation}
\end{equation}
where $\rho_{\mathrm{LR}}$ and $\rho_{\mathrm{HR}}$ denote the input and output sample densities of the SR model, respectively. For focal-length enlargement, setting $\rho_{\mathrm{LR}}=\rho_t$ and $s=s_f$ gives $\rho_{\mathrm{HR}}=s_f^2\rho_t=\rho_r$. Therefore, focal-length enlargement and SR share the same detail-recovery requirement under spatial magnification. In our pipeline, the scene is rendered directly under the target camera, and the SR prior restores the resulting coarse rendering by recovering details missing between sparsely projected samples. Viewpoint changes additionally reveal regions that are occluded or unobserved in the reference image, extending the task from detail restoration to geometry-aware completion. We therefore adapt the one-step TSD-SR model~\cite{dong2025tsdsr} as our generative backbone to address both requirements.
The adapted generator is jointly conditioned on the pixel-space rendering $\widetilde I_t$ and the feature-space projected tokens $\hat F_t$. Let $\mathcal{E}_{\mathrm{vae}}$ and $\mathcal{D}_{\mathrm{vae}}$ denote the VAE encoder and decoder, respectively, and let $\tau$ be the fixed denoising timestep. The dual-space-conditioned one-step generation process is defined as
\begin{equation}
    z_t=\mathcal{E}_{\mathrm{vae}}(\widetilde I_t),
    \qquad
    \hat I_t
    =
    \mathcal{D}_{\mathrm{vae}}
    \left(
    z_t-\epsilon_\theta(z_t,\tau;\hat F_t)
    \right).
    \label{eq:dual_conditioned_generation}
\end{equation}
We take the latent code of the pixel-space rendering $\widetilde I_t$ as the degraded input latent $z_t$ at the fixed timestep $\tau$ and use the adapted generative backbone to predict and remove the degradation residual in a single denoising step. The feature-space condition $\hat F_t$ is injected into the transformer attention layers to provide robust target-view guidance beyond the incomplete and locally misaligned rendering.

\begin{figure*}[!t]
    \centering
    \includegraphics[width=\textwidth]{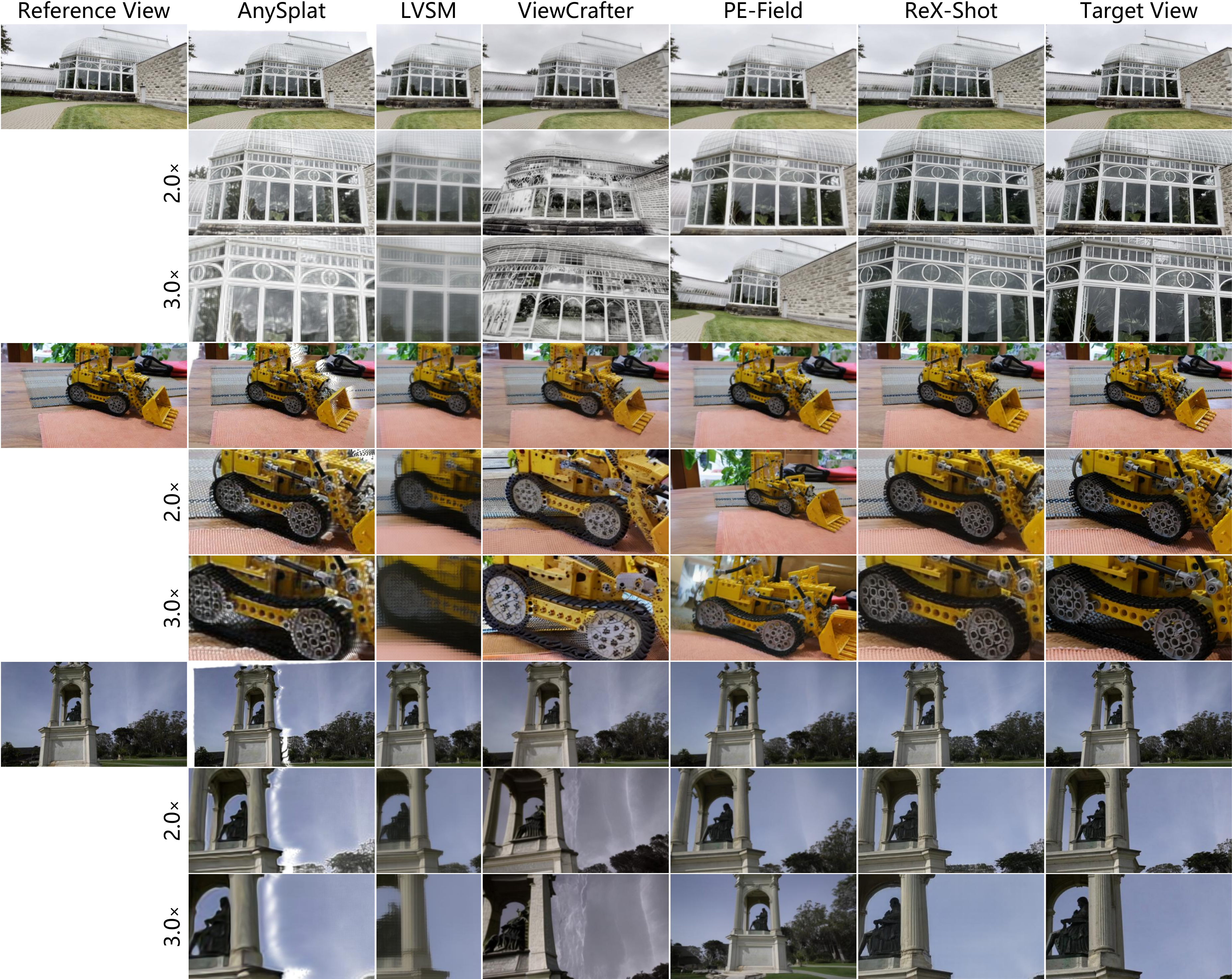}
    \caption{Qualitative comparison with novel-view-synthesis baselines under viewpoint and focal-length control. For each example, the first row shows results at the target viewpoint with the focal-length scale fixed at $1.0\times$. The second and third rows keep the target viewpoint fixed and enlarge the focal length by $2.0\times$ and $3.0\times$, respectively. The input reference and ground-truth target images are shown in the leftmost and rightmost columns, respectively.}
    \label{fig:nvs_focal}
\end{figure*}

Specifically, $\hat F_t$ is injected through an additional attention branch in every transformer layer. In parallel to the original attention output $A_\ell^{\mathrm{orig}}$, the feature tokens are normalized and linearly mapped to additional keys and values:
\begin{equation}
    \begin{aligned}
        k_\ell^F
        &= W_{k,\ell}^{F}\!\operatorname{Norm}(\hat F_t),
        \quad
        v_\ell^F
        = W_{v,\ell}^{F}\!\operatorname{Norm}(\hat F_t),\\
        A_\ell^{F}
        &=\operatorname{Attn}
        \left(
        q_\ell^h,
        [k_\ell^h;k_\ell^F],
        [v_\ell^h;v_\ell^F]
        \right),
    \end{aligned}
    \label{eq:dit_feature_injection}
\end{equation}
where $q_\ell^h$, $k_\ell^h$, and $v_\ell^h$ denote the query, key, and value corresponding to the image-latent tokens, while $k_\ell^F$ and $v_\ell^F$ are the feature keys and values derived from $\hat F_t$, respectively. The feature keys and values are concatenated with the image keys and values, respectively, along the token dimension before attention. The two attention outputs are combined as
\begin{equation}
    \widetilde A_\ell
    =
    A_\ell^{\mathrm{orig}}+A_\ell^{F}.
    \label{eq:dit_feature_fusion}
\end{equation}
This additional attention branch allows the image-latent tokens to interact directly with the projected foundation-model feature tokens at every transformer layer, while retaining the original attention pathway to preserve the pretrained generative prior.

During training, we fine-tune the TSD-SR-initialized generator with the paired target image $I_t$ as supervision. We combine mean squared error and LPIPS perceptual distance~\cite{zhang2018unreasonable} with equal weights:
\begin{equation}
    \mathcal{L}_{\mathrm{rephoto}}
    =\mathcal{L}_{\mathrm{MSE}}+\mathcal{L}_{\mathrm{LPIPS}}.
    \label{eq:rephotography_loss}
\end{equation}
This objective balances pixel-level fidelity and perceptual quality.

\subsection{Parameterized Photographic-Effect Control}
\label{sec:effect_control}
Alongside camera geometry, photographic effects are key elements of photography that govern image appearance. We therefore incorporate parameterized photographic-effect control into the 3D-aware generative backbone introduced above, enabling the model to preserve content consistency across viewpoints and focal lengths. For each effect type $\phi$, the corresponding photographic-effect encoder consists of a positional encoding MLP (PE-MLP) and a ControlNet-style auxiliary branch~\cite{Zhang_2023_ICCV} comprising DiT blocks~\cite{esser2024scaling} with the same architecture as those in the generative backbone. The PE-MLP maps the normalized control value $\alpha$ to an effect embedding $z_e$ using Fourier features~\cite{tancik2020fourier}:
\begin{equation}
    \begin{aligned}
    \gamma(\alpha)
    &=
    \left[
    \alpha,\{\sin(\omega_i \alpha),\cos(\omega_i \alpha)\}_{i=1}^{N_f}
    \right],\\
    z_e
    &=
    M_\phi\!\left(\gamma(\alpha)\right),
    \end{aligned}
    \label{eq:effect_encoding}
\end{equation}
where $M_\phi$ denotes the PE-MLP and $\{\omega_i\}_{i=1}^{N_f}$ denotes a set of fixed multi-scale frequencies. The Fourier encoding provides a multi-scale representation of $\alpha$, allowing the PE-MLP to distinguish nearby control values while modeling smooth, nonlinear effect variations. We then tile the resulting effect embedding over the latent spatial grid and apply patch embedding to obtain the initial effect tokens $g_0^e\in\mathbb{R}^{N_e\times d_h}$, with $N_e$ and $d_h$ matching the token count and hidden dimension of the main DiT branch, respectively.

The photographic-effect encoder is coupled with the main DiT layer by layer. Starting from $g_0^e$, its $\ell$-th DiT block produces a guidance signal $g_\ell^e$, which is injected after the corresponding main DiT block:
\begin{equation}
    \begin{aligned}
    g_\ell^e
    &=
    \operatorname{DiTBlock}_{\ell}^{e}(g_{\ell-1}^e),\\
    h_\ell
    &=
    \operatorname{DiTBlock}_{\ell}(h_{\ell-1})
    +g_\ell^e,
    \qquad \ell=1,\ldots,L_e.
    \end{aligned}
    \label{eq:effect_dit_injection}
\end{equation}
We train separate photographic-effect encoders for color temperature, exposure, saturation, and contrast using the reconstruction objective in Eq.~\eqref{eq:rephotography_loss} on the corresponding effect-conditioned target images. Throughout this stage, we freeze the main generative backbone and optimize only the corresponding photographic-effect encoder.

In this way, camera geometry (viewpoint and focal length) is handled by the main generative backbone, while the photographic-effect encoder leverages this 3D-aware backbone to maintain consistent effect control across viewpoints and focal lengths.

\section{EXPERIMENTS}
\label{sec:experiments}

We evaluate ReX-Shot from five perspectives: camera-geometry control, photographic-effect control, joint control, inference efficiency, and component contributions. We first describe the implementation details in Section~\ref{sec:implementation_details} and the experimental protocols in Section~\ref{sec:experimental_protocols}. We then evaluate viewpoint and focal-length control in Section~\ref{sec:view_focal_control}, followed by photographic-effect and joint control in Section~\ref{sec:effect_joint_control}. Finally, Section~\ref{sec:inference_efficiency} reports inference latency, and Section~\ref{sec:ablation_study} analyzes the contribution of the proposed components.

\subsection{Implementation Details}
\label{sec:implementation_details}
\noindent\textbf{Dataset.} We randomly split the 140 scenes of DL3DV-140~\cite{Ling_2024_CVPR} into 112 scenes for training and 28 scenes for testing. For each training scene, we group every nine consecutive frames into a cluster, obtaining 4,415 clusters in total. Within each cluster, the center frame serves as the reference input, while each of the other eight frames provides a target view, resulting in 35,320 reference-target pairs. To simulate focal-length changes, we independently sample the focal-length scale $s_f$ for each target from the continuous interval $[1.0, 3.0]$ and generate the corresponding ground truth via center cropping. For photographic-effect training, we sample color temperature from $[2000, 10000]$\,K, exposure scale from $[0.2, 2.0]$, saturation scale from $[0.0, 2.0]$, and contrast scale from $[0.5, 2.0]$. Each sampled value is normalized to $[0,1]$ using its corresponding control range before being passed to the photographic-effect encoder. We evaluate the model on the DL3DV-140 test split and additionally on MipNeRF360~\cite{Barron_2022_CVPR} and Tanks \& Temples~\cite{Knapitsch2017} to assess cross-dataset generalization.

\noindent\textbf{Training details.} We implement ReX-Shot on top of Stable Diffusion 3 Medium~\cite{esser2024scaling} and initialize the generative backbone from the released TSD-SR~\cite{dong2025tsdsr} checkpoint. We begin by pretraining the camera-aware feature projector using frozen VGGT~\cite{wang2025vggt} and DINOv2~\cite{oquab2023dinov2} feature extractors. With the pretrained projector and the TSD-SR LoRA weights as initialization, we then fine-tune the generative backbone on the DL3DV-140 training split by jointly optimizing LoRA adapters~\cite{hu2022lora} for the SD3 transformer, VAE encoder, and projector branch, along with the feature-injection attention processors. This fine-tuning stage uses 4 GPUs with a batch size of 8 per GPU, a learning rate of $1\times10^{-4}$, 500 warmup steps, 8 epochs, and a training resolution of $448\times252$ pixels. After completing this stage, we freeze the SD3 transformer, VAE, and camera-aware feature projector, and train separate photographic-effect encoders, one for each effect. These encoders use the same training resolution and optimization settings as the preceding generative-backbone fine-tuning stage.

\begin{table*}[t]
    \centering
    \caption{Quantitative comparison under joint viewpoint and focal-length control.}
    \label{tab:exp_nvs_focal}
    \begin{tabularx}{0.95\textwidth}{l*{12}{C}}
        \toprule
        \multirow{2}{*}{Method} & \multicolumn{3}{c}{DL3DV} & \multicolumn{3}{c}{MipNeRF360} & \multicolumn{3}{c}{Tanks \& Temples} & \multicolumn{3}{c}{Average} \\
        \cmidrule(lr){2-4}\cmidrule(lr){5-7}\cmidrule(lr){8-10}\cmidrule(lr){11-13}
        & PSNR$\uparrow$ & SSIM$\uparrow$ & LPIPS$\downarrow$ & PSNR$\uparrow$ & SSIM$\uparrow$ & LPIPS$\downarrow$ & PSNR$\uparrow$ & SSIM$\uparrow$ & LPIPS$\downarrow$ & PSNR$\uparrow$ & SSIM$\uparrow$ & LPIPS$\downarrow$ \\
        \midrule
        AnySplat          &          \third{16.51}          &          \third{0.464}          &          0.461          &          14.66          &          \third{0.457}          &          \third{0.477}          &          13.61          &          0.425          &          0.521          &          14.93          &          \third{0.449}          &          0.486 \\
        CameraCtrl          &          14.45          &          0.391          &          0.480          &          13.71          &          0.400          &          0.604          &          12.69          &          0.411          &          0.545          &          13.62          &          0.401          &          0.543 \\
        LVSM          &          \second{18.87}          &          \second{0.526}          &          \second{0.383}          &          15.48          &          \second{0.498}          &          0.636          &          \second{15.62}          &          \second{0.514}          &          0.511          &          \second{16.66}          &          \second{0.513}          &          0.510 \\
        ViewCrafter          &          15.84          &          0.417          &          \third{0.424}          &          \second{16.13}          &          0.443          &          \second{0.453}          &          14.18          &          0.433          &          \third{0.487}          &          15.38          &          0.431          &          \second{0.455} \\
        PE-Field          &          16.34          &          0.429          &          0.427          &          \third{15.95}          &          0.443          &          0.486          &          \third{14.31}          &          \third{0.433}          &          \second{0.485}          &          \third{15.53}          &          0.435          &          \third{0.466} \\
        ReX-Shot          &          \best{23.05}          &          \best{0.679}          &          \best{0.137}          &          \best{19.93}          &          \best{0.557}          &          \best{0.276}          &          \best{18.07}          &          \best{0.533}          &          \best{0.275}          &          \best{20.35}          &          \best{0.590}          &          \best{0.229} \\
        \bottomrule
    \end{tabularx}
\end{table*}

\begin{table*}[t]
    \centering
    \caption{Quantitative comparison under viewpoint-only control.}
    \label{tab:exp_nvs}
    \begin{tabularx}{0.95\textwidth}{l*{12}{C}}
        \toprule
        \multirow{2}{*}{Method} & \multicolumn{3}{c}{DL3DV} & \multicolumn{3}{c}{MipNeRF360} & \multicolumn{3}{c}{Tanks \& Temples} & \multicolumn{3}{c}{Average} \\
        \cmidrule(lr){2-4}\cmidrule(lr){5-7}\cmidrule(lr){8-10}\cmidrule(lr){11-13}
        & PSNR$\uparrow$ & SSIM$\uparrow$ & LPIPS$\downarrow$ & PSNR$\uparrow$ & SSIM$\uparrow$ & LPIPS$\downarrow$ & PSNR$\uparrow$ & SSIM$\uparrow$ & LPIPS$\downarrow$ & PSNR$\uparrow$ & SSIM$\uparrow$ & LPIPS$\downarrow$ \\
        \midrule
        AnySplat        &        18.07        &        \second{0.638}        &        0.223        &        12.39        &        \third{0.479}        &        0.419        &        12.65        &        \second{0.526}        &        \third{0.386}        &        14.37        &        \second{0.548}        &        0.343 \\
        CameraCtrl        &        17.63        &        0.481        &        0.273        &        15.56        &        0.438        &        0.503        &        14.53        &        0.469        &        0.428        &        15.91        &        0.463        &        0.401 \\
        LVSM        &        \third{19.82}        &        0.564        &        \third{0.210}        &        15.74        &        0.459        &        0.523        &        \third{15.86}        &        \third{0.510}        &        0.388        &        17.14        &        \third{0.511}        &        0.374 \\
        ViewCrafter        &        \second{19.94}        &        \third{0.570}        &        \second{0.205}        &        \third{17.37}        &        0.462        &        \third{0.380}        &        15.30        &        0.470        &        0.397        &        \third{17.54}        &        0.501        &        \third{0.327} \\
        PE-Field        &        19.55        &        0.510        &        0.236        &        \second{18.59}        &        \second{0.491}        &        \second{0.344}        &        \second{16.49}        &        0.495        &        \second{0.348}        &        \second{18.21}        &        0.499        &        \second{0.309} \\
        ReX-Shot        &        \best{23.13}        &        \best{0.703}        &        \best{0.122}        &        \best{20.18}        &        \best{0.548}        &        \best{0.245}        &        \best{17.87}        &        \best{0.544}        &        \best{0.244}        &        \best{20.39}        &        \best{0.598}        &        \best{0.204} \\
        \bottomrule
    \end{tabularx}
\end{table*}

\begin{table*}[t]
    \centering
    \caption{Quantitative comparison under focal-length-only control.}
    \label{tab:exp_focal}
    \begin{tabularx}{0.95\textwidth}{l*{12}{C}}
        \toprule
        \multirow{2}{*}{Method} & \multicolumn{3}{c}{DL3DV} & \multicolumn{3}{c}{MipNeRF360} & \multicolumn{3}{c}{Tanks \& Temples} & \multicolumn{3}{c}{Average} \\
        \cmidrule(lr){2-4}\cmidrule(lr){5-7}\cmidrule(lr){8-10}\cmidrule(lr){11-13}
        & PSNR$\uparrow$ & SSIM$\uparrow$ & LPIPS$\downarrow$ & PSNR$\uparrow$ & SSIM$\uparrow$ & LPIPS$\downarrow$ & PSNR$\uparrow$ & SSIM$\uparrow$ & LPIPS$\downarrow$ & PSNR$\uparrow$ & SSIM$\uparrow$ & LPIPS$\downarrow$ \\
        \midrule
        AnySplat      &      20.21      &      0.676      &      0.398      &      \third{25.39}      &      \second{0.797}      &      0.285      &      20.90      &      0.695      &      0.365      &      22.17      &      \second{0.723}      &      0.349 \\
        CameraCtrl      &      15.65      &      0.430      &      0.409      &      15.20      &      0.426      &      0.488      &      14.50      &      0.448      &      0.433      &      15.12      &      0.435      &      0.443 \\
        LVSM      &      22.16      &      0.628      &      0.259      &      24.37      &      0.689      &      0.279      &      22.98      &      0.681      &      0.247      &      23.17      &      0.666      &      0.262 \\
        ViewCrafter      &      16.40      &      0.428      &      0.427      &      19.70      &      0.551      &      0.403      &      15.90      &      0.486      &      0.408      &      17.33      &      0.488      &      0.413 \\
        PE-Field      &      17.08      &      0.453      &      0.390      &      18.47      &      0.509      &      0.359      &      16.37      &      0.489      &      0.389      &      17.31      &      0.484      &      0.379 \\
        OSEDiff      &      \third{23.22}      &      \third{0.696}      &      \third{0.208}      &      24.39      &      0.698      &      \third{0.236}      &      \third{23.66}      &      \third{0.725}      &      \third{0.194}      &      \third{23.76}      &      0.706      &      \third{0.213} \\
        TSD-SR      &      \second{24.02}      &      \second{0.703}      &      \second{0.166}      &      \second{26.19}      &      \third{0.717}      &      \second{0.165}      &      \second{25.37}      &      \second{0.743}      &      \second{0.146}      &      \second{25.19}      &      \third{0.721}      &      \second{0.159} \\
        ReX-Shot      &      \best{30.00}      &      \best{0.878}      &      \best{0.058}      &      \best{28.57}      &      \best{0.846}      &      \best{0.128}      &      \best{28.83}      &      \best{0.859}      &      \best{0.098}      &      \best{29.13}      &      \best{0.861}      &      \best{0.095} \\
        \bottomrule
    \end{tabularx}
\end{table*}

\subsection{Experimental Protocols}
\label{sec:experimental_protocols}
\noindent\textbf{Evaluation settings.} Our evaluation covers camera-geometry control, photographic-effect control, joint control, inference efficiency, and component contributions. For camera-geometry control, we evaluate viewpoint and focal-length changes on DL3DV-140, MipNeRF360, and Tanks \& Temples using PSNR~\cite{huynh2008scope}, SSIM~\cite{wang2004image}, and LPIPS~\cite{zhang2018unreasonable}. For photographic-effect control, we evaluate color temperature, exposure, saturation, and contrast, and additionally report $\Delta E$~\cite{sharma2005ciede2000} to measure color accuracy. For joint control, we demonstrate ReX-Shot's ability to jointly control camera geometry and photographic effects within a single generation by simultaneously varying viewpoint, focal length, and color temperature. For inference efficiency, we measure end-to-end latency and decompose it into reconstruction, rendering, and effect-editing time where applicable. For component contributions, we conduct ablation studies to isolate the effects of the TSD-SR initialization and the proposed camera-aware feature projector. In each quantitative table, the best, second-best, and third-best results for each metric are highlighted with \colorswatch{bestcolor}, \colorswatch{secondcolor}, and \colorswatch{thirdcolor}, respectively.

\noindent\textbf{Baselines.} The baselines are selected according to the capability required by each protocol. For camera-geometry control, we compare with AnySplat~\cite{jiang2025anysplat}, LVSM~\cite{jin2025lvsm}, CameraCtrl~\cite{he2024cameractrl}, ViewCrafter~\cite{yu2025viewcrafter}, and PE-Field~\cite{bai2026pefield} to examine whether existing novel-view-synthesis methods remain reliable when the viewpoint and focal length change simultaneously. To disentangle these two factors, we also evaluate these baselines under a viewpoint-only setting and a focal-length-only setting. OSEDiff~\cite{wu2024osediff} and TSD-SR~\cite{dong2025tsdsr} are additionally included in the focal-length-only setting because focal-length enlargement poses a detail-recovery problem closely related to super resolution. For photographic-effect control, we compare with IP2P~\cite{Brooks_2023_CVPR} and ICEdit~\cite{zhang2025icedit} as general-purpose baselines because they support a broad range of appearance edits, and photographic effects can also be described using natural-language instructions. For inference efficiency, we compare ReX-Shot with representative feed-forward and generative methods. We use the official implementations and released checkpoints for all baselines with their recommended inference settings. All baselines are evaluated on the same reference-target pairs.

\subsection{Viewpoint and Focal Length Control}
\label{sec:view_focal_control}

We evaluate geometry control by varying viewpoint and focal length within the test-set clusters. Each cluster contains nine neighboring viewpoints, and each viewpoint is evaluated under nine uniformly sampled focal-length scales from $1.0\times$ to $3.0\times$. From these reference-target pairs, we derive three evaluation settings. The viewpoint-only protocol uses samples whose focal length remains at $1.0\times$ while the target viewpoint changes. The focal-length-only protocol keeps the viewpoint fixed to the reference view and evaluates the remaining focal-length scales. The joint viewpoint and focal length protocol uses the samples where both the target viewpoint and the focal length differ from those of the reference view. Together, these three protocols comprehensively evaluate viewpoint and focal-length control, both independently and in combination.

Fig.~\ref{fig:nvs_focal} evaluates whether state-of-the-art novel-view-synthesis methods can extend beyond standard viewpoint transfer to handle focal-length changes, thereby supporting comprehensive camera-geometry control. The first row of each example follows the viewpoint-only protocol, where the target viewpoint changes while the focal-length scale remains fixed at $1.0\times$. In this case, ReX-Shot achieves visual quality comparable to strong novel-view-synthesis baselines under the single-image setting. The second and third rows are more diagnostic, with the target viewpoint fixed to that in the first row while the focal length is enlarged by $2.0\times$ and $3.0\times$. AnySplat can render geometrically plausible content using its predicted explicit scene representation and camera parameters, but it cannot synthesize content in previously unobserved regions that become visible under viewpoint changes, including areas occluded or outside the reference field of view. Focal-length enlargement further produces visible gaps between projected primitives. LVSM, meanwhile, exhibits grid-like artifacts. ViewCrafter hallucinates unsupported textures and distorts object geometry, whereas PE-Field produces an incorrect field of view. These results highlight the limitations of existing novel-view-synthesis methods in handling focal-length changes for rephotography. In contrast, ReX-Shot plausibly synthesizes content in disoccluded and previously out-of-view regions under viewpoint changes and recovers fine-grained texture details under focal-length enlargement.

\begin{figure}[!t]
    \centering
    \includegraphics[width=\columnwidth]{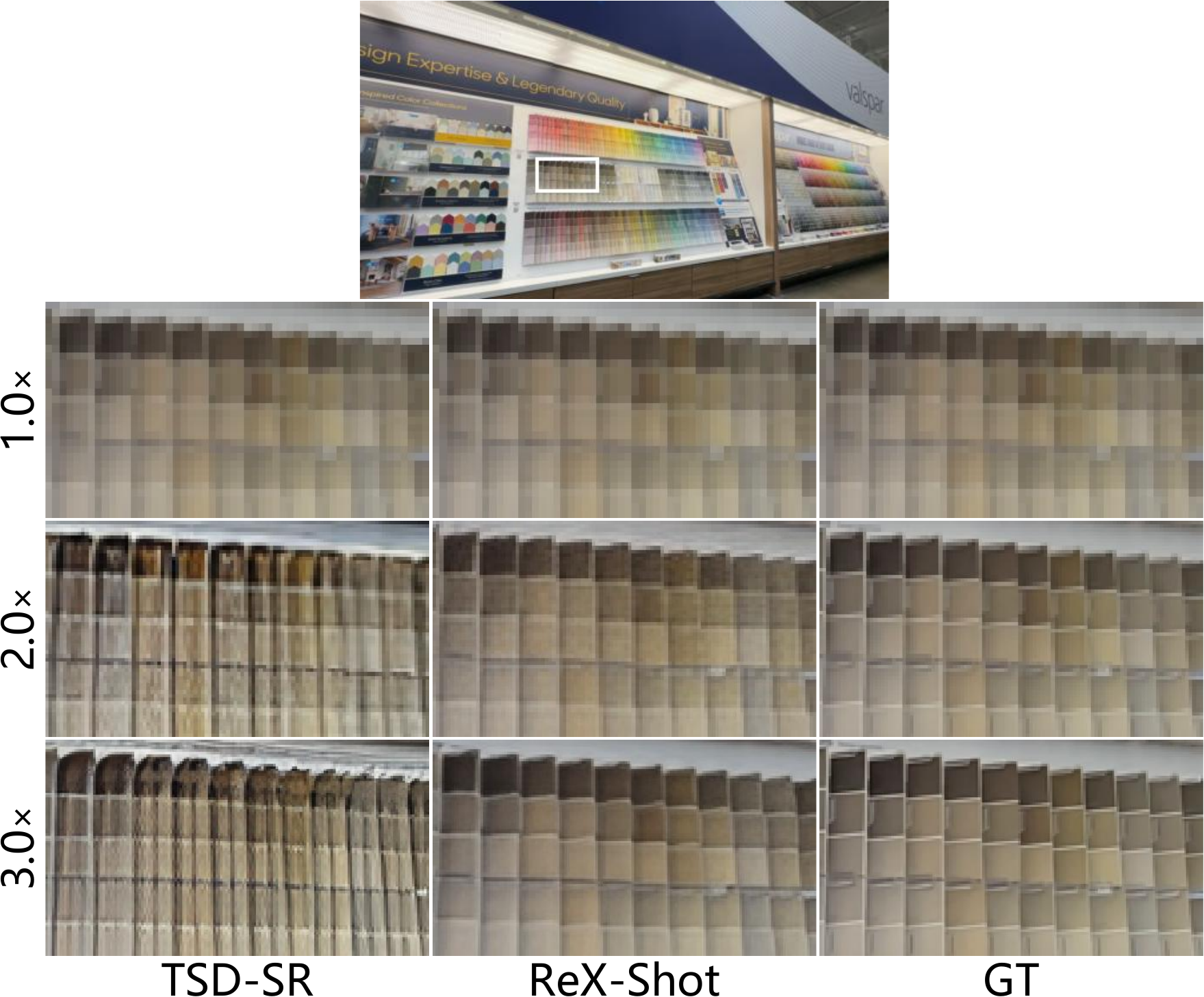}
    \caption{Qualitative focal-length-only comparison between TSD-SR and ReX-Shot under continuous focal-length enlargement.}
    \label{fig:focal}
\end{figure}

\begin{table}[t]
    \centering
    \caption{Quantitative comparison under photographic-effect control.}
    \label{tab:exp_effect}
    \begin{tabularx}{0.95\columnwidth}{ll*{3}{C}}
        \toprule
        Metric & Effect & ICEdit & IP2P & ReX-Shot \\
        \midrule
        \multirow{5}{*}{PSNR$\uparrow$}
            &     Color Temp.     &     \second{23.28}     &     \third{20.23}     &     \best{31.72} \\
            &     Exposure     &     \second{17.25}     &     \third{16.13}     &     \best{30.18} \\
            &     Saturation     &     \third{19.92}     &     \second{27.75}     &     \best{30.13} \\
            &     Contrast     &     \third{19.86}     &     \second{20.64}     &     \best{28.59} \\
            &     Average     &     \third{20.08}     &     \second{21.19}     &     \best{30.16} \\
        \midrule
        \multirow{5}{*}{SSIM$\uparrow$}
            &     Color Temp.     &     \second{0.909}     &     \third{0.781}     &     \best{0.913} \\
            &     Exposure     &     \second{0.791}     &     \third{0.697}     &     \best{0.907} \\
            &     Saturation     &     \third{0.814}     &     \second{0.832}     &     \best{0.905} \\
            &     Contrast     &     \second{0.791}     &     \third{0.759}     &     \best{0.897} \\
            &     Average     &     \second{0.826}     &     \third{0.767}     &     \best{0.906} \\
        \midrule
        \multirow{5}{*}{LPIPS$\downarrow$}
            &     Color Temp.     &     \second{0.199}     &     \third{0.270}     &     \best{0.152} \\
            &     Exposure     &     \best{0.166}     &     \third{0.203}     &     \best{0.166} \\
            &     Saturation     &     \third{0.293}     &     \best{0.129}     &     \second{0.169} \\
            &     Contrast     &     \third{0.212}     &     \second{0.193}     &     \best{0.163} \\
            &     Average     &     \third{0.218}     &     \second{0.199}     &     \best{0.163} \\
        \midrule
        \multirow{5}{*}{$\Delta E\downarrow$}
            &     Color Temp.     &     \second{13.71}     &     \third{18.23}     &     \best{3.033} \\
            &     Exposure     &     \second{17.92}     &     \third{18.00}     &     \best{3.814} \\
            &     Saturation     &     \third{17.07}     &     \second{7.092}     &     \best{3.715} \\
            &     Contrast     &     \second{11.79}     &     \third{12.16}     &     \best{4.099} \\
            &     Average     &     \third{15.12}     &     \second{13.87}     &     \best{3.665} \\
        \bottomrule
    \end{tabularx}
\end{table}

\begin{figure*}[!t]
    \centering
    \includegraphics[width=\textwidth]{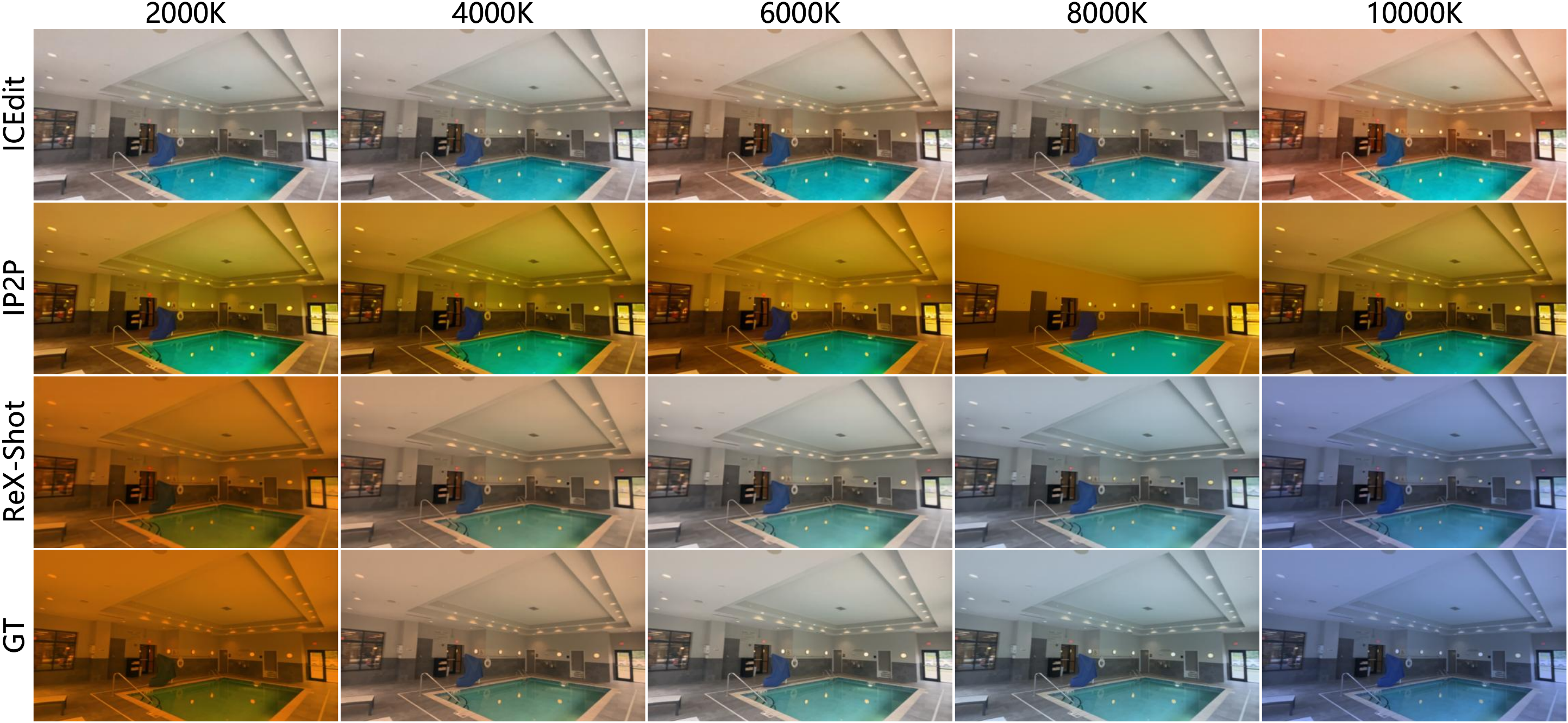}
    \caption{Qualitative comparison with instruction-guided image editing baselines under color-temperature control.}
    \label{fig:color_temperature}
\end{figure*}

\begin{figure*}[!t]
    \centering
    \includegraphics[width=\textwidth]{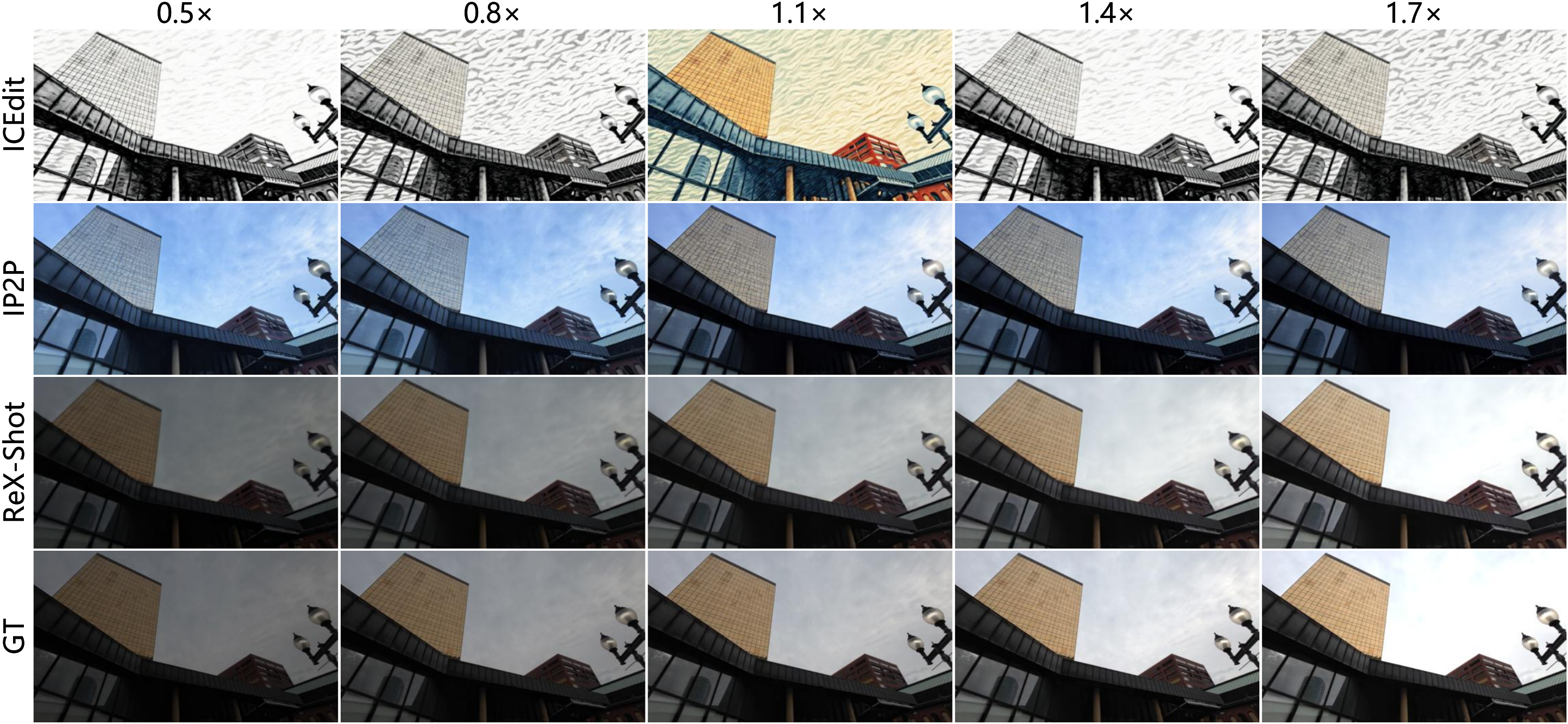}
    \caption{Qualitative comparison with instruction-guided image editing baselines under exposure control.}
    \label{fig:exposure}
\end{figure*}

\begin{figure*}[!t]
    \centering
    \includegraphics[width=\textwidth]{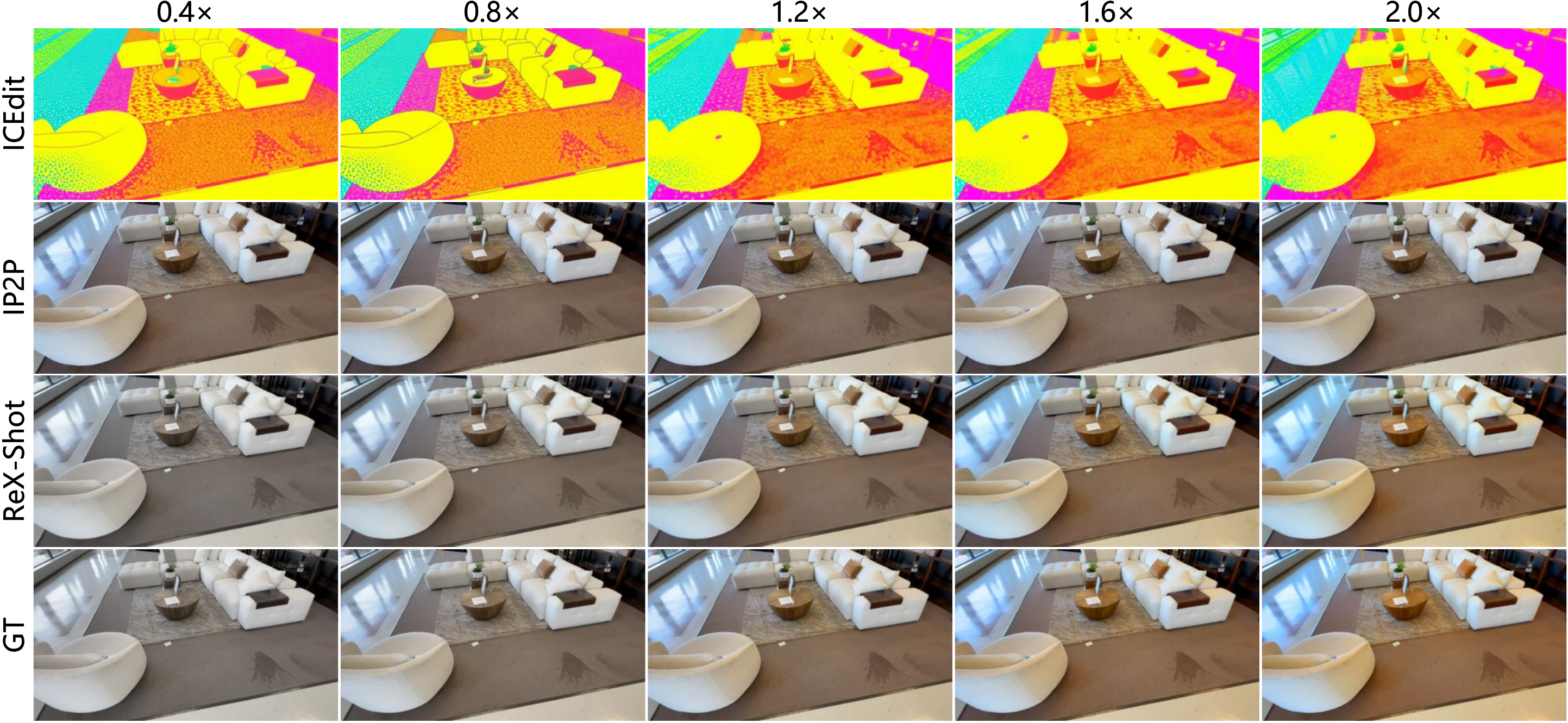}
    \caption{Qualitative comparison with instruction-guided image editing baselines under saturation control.}
    \label{fig:saturation}
\end{figure*}

\begin{figure*}[!t]
    \centering
    \includegraphics[width=\textwidth]{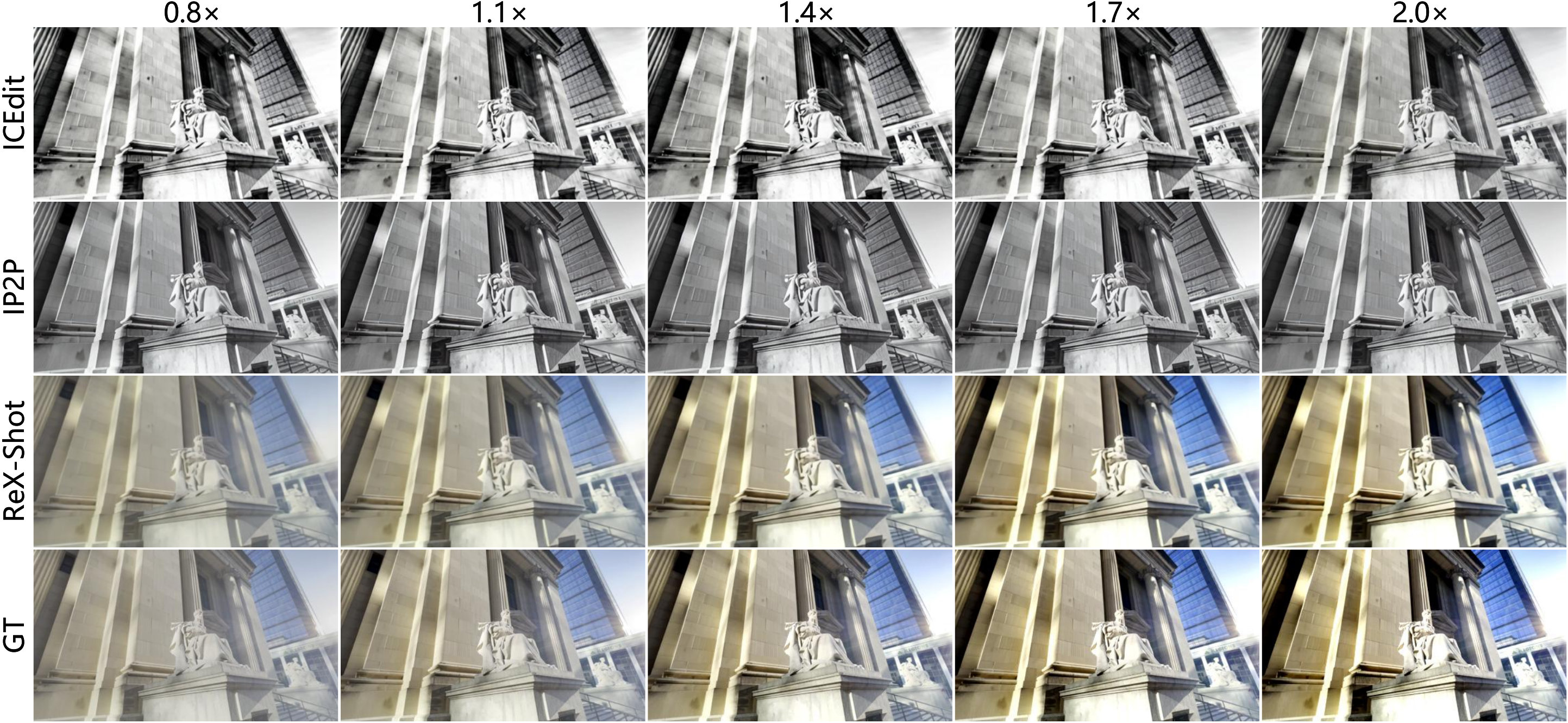}
    \caption{Qualitative comparison with instruction-guided image editing baselines under contrast control.}
    \label{fig:contrast}
\end{figure*}

The quantitative results in Tab.~\ref{tab:exp_nvs_focal} and Tab.~\ref{tab:exp_nvs} reinforce these qualitative observations. Under joint viewpoint and focal-length control, ReX-Shot outperforms all compared methods across the three datasets, demonstrating that correct focal-length control cannot be obtained by directly applying novel-view-synthesis baselines. In the viewpoint-only setting, ReX-Shot remains comparable to strong baselines such as PE-Field, ViewCrafter, and LVSM, indicating that ReX-Shot retains strong novel-view-synthesis performance while supporting focal-length control.

The focal-length-only experiment isolates focal-length control while keeping the camera viewpoint unchanged. This setting is evaluated against all novel-view-synthesis baselines as well as generative super-resolution baselines, with particular attention to OSEDiff and TSD-SR, as they deliver strong results in Tab.~\ref{tab:exp_focal}. Nevertheless, a successful method should not only recover fine-grained details during focal-length enlargement, but also preserve scene geometry and maintain consistency with the reference observation across continuous focal-length changes. We therefore select TSD-SR as a representative method and compare it with ReX-Shot in Fig.~\ref{fig:focal}. TSD-SR treats focal-length enlargement as a 2D restoration problem. Its strong generative prior can enhance fine details, but without explicit geometric guidance, it may hallucinate implausible textures and compromise geometric consistency. ReX-Shot instead provides geometry- and camera-grounded guidance for focal-length control, allowing its generative prior to recover fine details while better preserving geometric consistency and scene fidelity. This advantage is reflected in the best PSNR, SSIM, and LPIPS in Tab.~\ref{tab:exp_focal}.

\subsection{Photographic-Effect and Joint Control}
\label{sec:effect_joint_control}

We evaluate parameterized photographic-effect control against IP2P and ICEdit. The two baselines edit images through natural-language instructions. For both methods, we use the prompt \texttt{adjust the \{photographic\_effect\} of this image to \{effect\_value\}}, replacing the placeholders with the target effect type and value. At test time, we uniformly sample several target values from 2000\,K to 10000\,K for color temperature, from $0.2\times$ to $2.0\times$ for exposure, from $0.0\times$ to $2.0\times$ for saturation, and from $0.5\times$ to $2.0\times$ for contrast.

The qualitative results in Figs.~\ref{fig:color_temperature}--\ref{fig:contrast} reveal two observations. First, their responses to numerical instructions are unstable, with effect strengths often poorly aligned with the requested values and some inputs leading to severe failure cases, as shown in the first row of Fig.~\ref{fig:saturation}. Second, the edits may alter the underlying scene content instead of changing only the photographic effect, as illustrated by the unintended changes to the ceiling in the fourth column of the second row in Fig.~\ref{fig:color_temperature} and to the sky in the first row of Fig.~\ref{fig:exposure}. ReX-Shot instead anchors scene content with deterministic geometry and introduces a parameterized photographic-effect encoder for continuous effect control. This design decouples scene content from photographic-effect control, avoiding unintended content changes while enabling accurate and continuous effect control for rephotography. Tab.~\ref{tab:exp_effect} confirms the visual trend, with ReX-Shot achieving the best average PSNR, SSIM, and LPIPS, as well as a substantially lower average $\Delta E$ than both IP2P and ICEdit.

\begin{figure}[!t]
    \centering
    \includegraphics[width=\columnwidth]{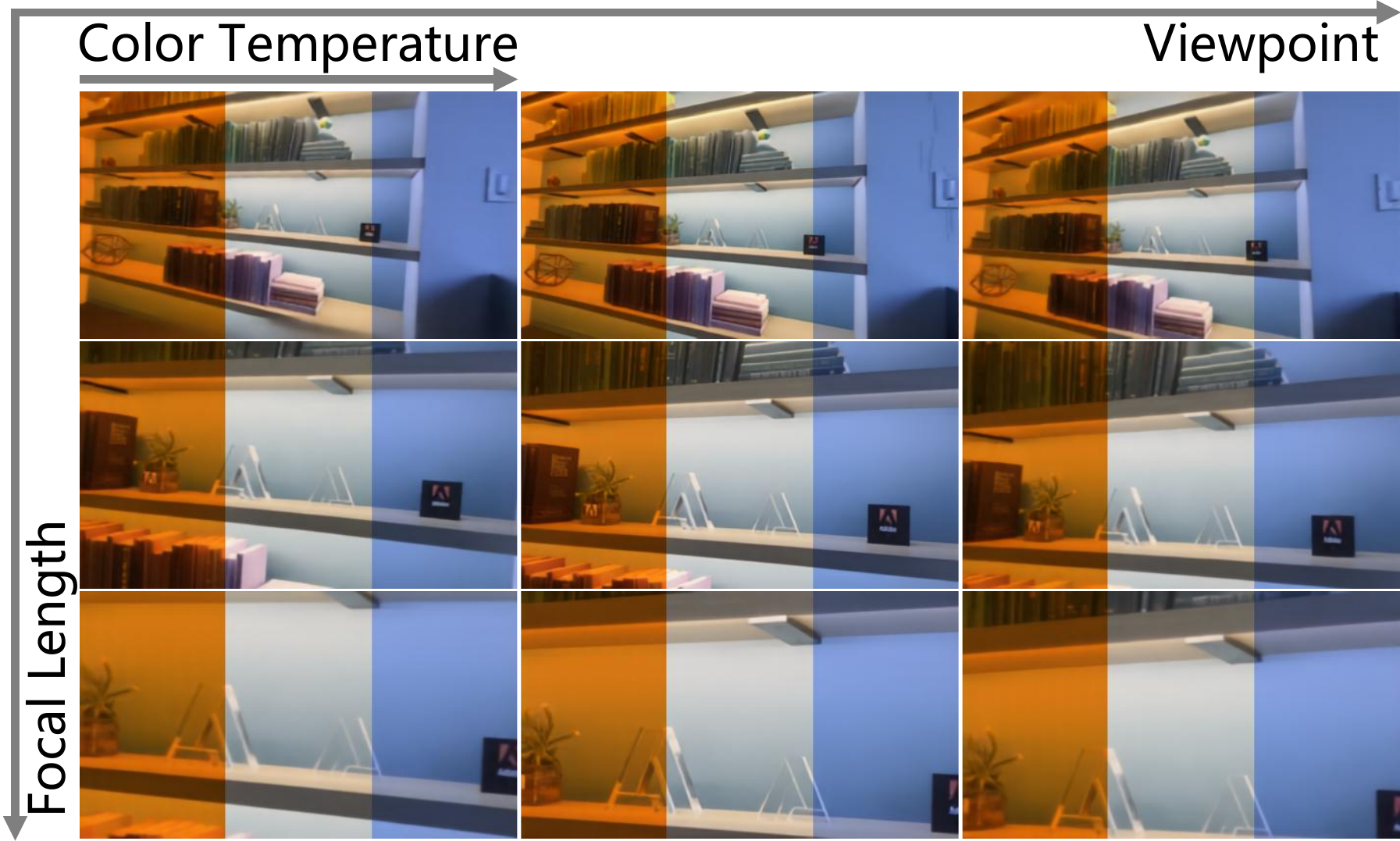}
    \caption{Qualitative results under joint control of viewpoint, focal length, and color temperature.}
    \label{fig:combination}
\end{figure}

Fig.~\ref{fig:combination} further presents an example of joint camera-geometry and photographic-effect control. Rather than applying these controls through independent post-processing steps, ReX-Shot realizes them jointly within a unified generation process. Enabled by its 3D-aware generative backbone and parameterized photographic-effect encoder, ReX-Shot produces geometrically consistent results across viewpoints and focal lengths while faithfully realizing the requested photographic effect.

\subsection{Inference Efficiency}
\label{sec:inference_efficiency}

\begin{figure*}[!t]
    \centering
    \includegraphics[width=\textwidth]{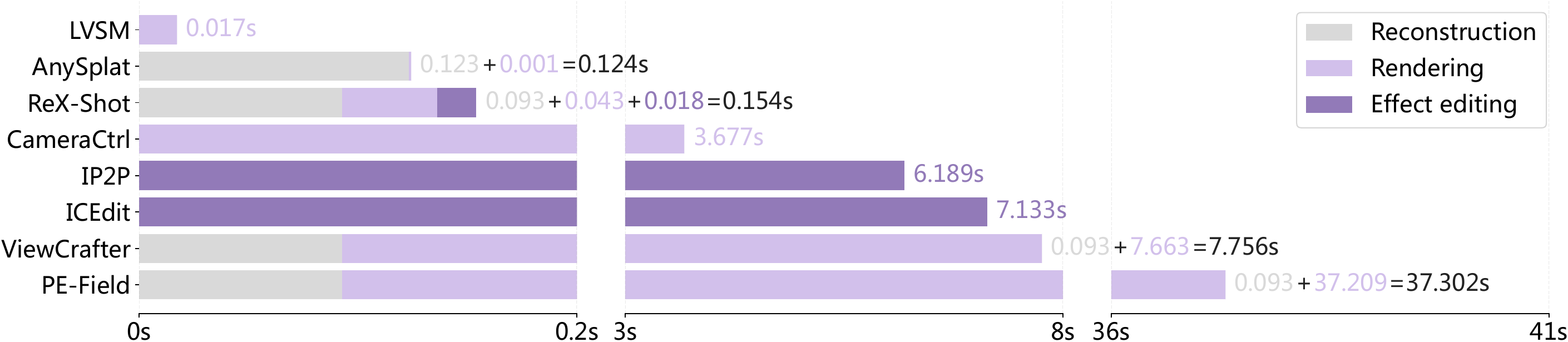}
    \caption{Inference latency comparison between ReX-Shot and representative feed-forward and generative methods. When applicable, end-to-end latency is decomposed into reconstruction, rendering for camera-geometry control, and effect editing for photographic-effect control.}
    \label{fig:timing_comparison}
\end{figure*}

Fig.~\ref{fig:timing_comparison} compares the end-to-end inference latency of ReX-Shot with feed-forward methods (LVSM and AnySplat) and generative methods (CameraCtrl, IP2P, ICEdit, ViewCrafter, and PE-Field). Where applicable, we decompose the end-to-end latency into reconstruction, target-camera rendering, and photographic-effect editing. Notably, although LVSM~\cite{jin2025lvsm} and CameraCtrl~\cite{he2024cameractrl} do not directly rely on 3D reconstruction in their pipelines, their camera control still requires calibrated camera parameters, which are commonly estimated together with scene structure by structure-from-motion pipelines~\cite{Schonberger_2016_CVPR}. Leveraging the efficiency of its one-step generative backbone, ReX-Shot achieves end-to-end inference latency comparable to feed-forward methods and substantially lower than that of other generative methods, while jointly supporting viewpoint, focal-length, and photographic-effect control. Moreover, after a one-time reconstruction of the reference image, the resulting scene representation can be reused across target camera and effect settings, so subsequent outputs require only rendering and effect control. Rendering alone runs at approximately 23 FPS, while rephotography with photographic-effect control runs at approximately 16 FPS, enabling near-real-time interaction.

\begin{table*}[t]
    \centering
    \caption{Quantitative comparison across SD3-FT, TSD-SR-FT, and TSD-SR-FT+FP. Parentheses report changes relative to the preceding row.}
    \label{tab:ablation}
    \setlength{\tabcolsep}{0.5pt}
    \newlength{\abgainwidth}
    \settowidth{\abgainwidth}{\,{\fontsize{6}{7}\selectfont (+0.000)}}
    \newcommand{\abbase}[1]{#1\makebox[\abgainwidth][l]{}}
    \newcommand{\abgain}[2]{#1\makebox[\abgainwidth][l]{\,{\fontsize{6}{7}\selectfont (#2)}}}
    \begin{tabularx}{0.95\textwidth}{l*{9}{C}}
        \toprule
        \multirow{2}{*}{Method} & \multicolumn{3}{c}{Focal Length Only} & \multicolumn{3}{c}{Viewpoint Only} & \multicolumn{3}{c}{Joint Control} \\
        \cmidrule(lr){2-4}\cmidrule(lr){5-7}\cmidrule(lr){8-10}
        & PSNR$\uparrow$ & SSIM$\uparrow$ & LPIPS$\downarrow$ & PSNR$\uparrow$ & SSIM$\uparrow$ & LPIPS$\downarrow$ & PSNR$\uparrow$ & SSIM$\uparrow$ & LPIPS$\downarrow$ \\
        \midrule
        SD3-FT    &    \third{\abbase{25.42}}    &    \third{\abbase{0.764}}    &    \third{\abbase{0.141}}    &    \third{\abbase{20.82}}    &    \third{\abbase{0.590}}    &    \third{\abbase{0.216}}    &    \third{\abbase{20.69}}    &    \third{\abbase{0.566}}    &    \third{\abbase{0.230}} \\
        TSD-SR-FT    &    \second{\abgain{29.64}{+4.22}}    &    \second{\abgain{0.867}{+0.103}}    &    \second{\abgain{0.061}{\textminus0.080}}    &    \second{\abgain{22.30}{+1.48}}    &    \second{\abgain{0.658}{+0.068}}    &    \second{\abgain{0.134}{\textminus0.082}}    &    \second{\abgain{21.85}{+1.16}}    &    \second{\abgain{0.620}{+0.054}}    &    \second{\abgain{0.153}{\textminus0.077}} \\
        TSD-SR-FT+FP    &    \best{\abgain{30.00}{+0.36}}    &    \best{\abgain{0.878}{+0.011}}    &    \best{\abgain{0.058}{\textminus0.003}}    &    \best{\abgain{23.13}{+0.83}}    &    \best{\abgain{0.703}{+0.045}}    &    \best{\abgain{0.122}{\textminus0.012}}    &    \best{\abgain{23.05}{+1.20}}    &    \best{\abgain{0.679}{+0.059}}    &    \best{\abgain{0.137}{\textminus0.016}} \\
        \bottomrule
    \end{tabularx}
\end{table*}

\begin{figure}[!t]
    \centering
    \includegraphics[width=\columnwidth]{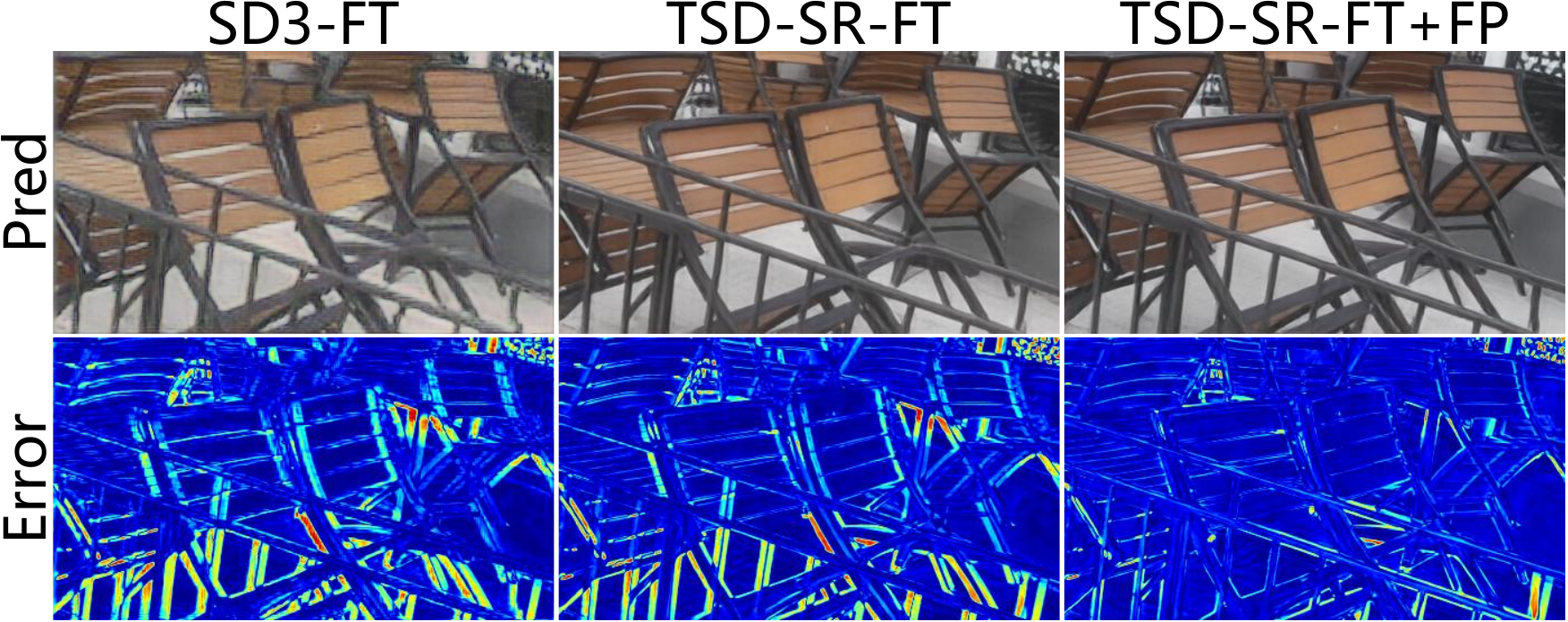}
    \caption{Qualitative comparison across SD3-FT, TSD-SR-FT, and TSD-SR-FT+FP. The top row shows the generated images, and the bottom row shows the corresponding error maps. All error maps are visualized using the same color scale, with warmer colors indicating larger errors.}
    \label{fig:ablation}
\end{figure}

\begin{figure}[!t]
    \centering
    \includegraphics[width=\columnwidth]{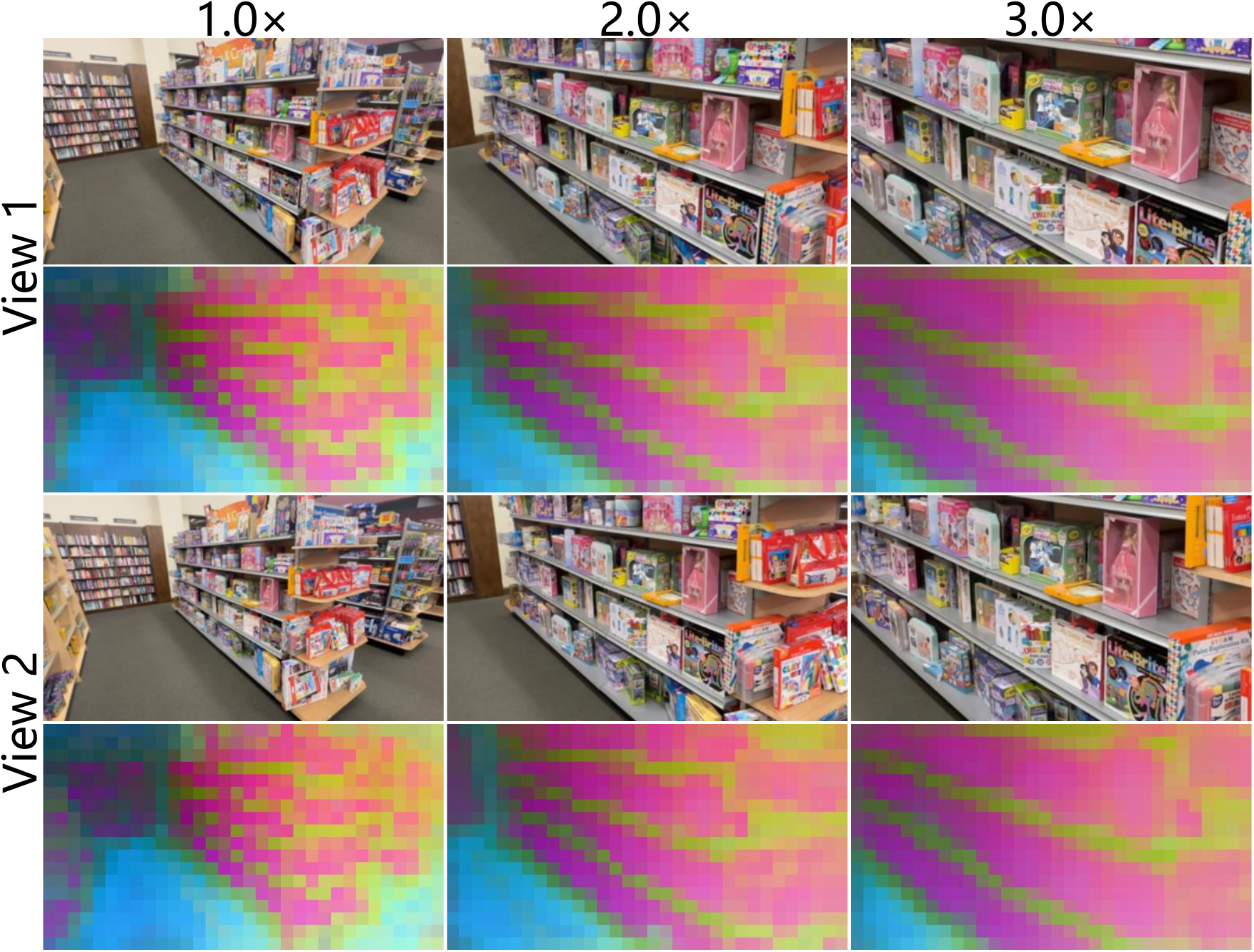}
    \caption{PCA visualization of the projected output features, with the first three principal components mapped to RGB. The two rows correspond to two target viewpoints, and the columns show focal-length scales of $1.0\times$, $2.0\times$, and $3.0\times$.}
    \label{fig:feature_projection_module}
\end{figure}

\subsection{Ablation Study}
\label{sec:ablation_study}

The ablation study evaluates the contributions of the super-resolution prior and the camera-aware feature projector. Fig.~\ref{fig:ablation} compares SD3 fine-tuning (SD3-FT), TSD-SR fine-tuning (TSD-SR-FT), and TSD-SR fine-tuning with the camera-aware feature projector (TSD-SR-FT+FP). SD3-FT often produces blurry and noisy results with noticeable local offsets. TSD-SR-FT improves detail recovery and perceptual quality, demonstrating the benefit of the super-resolution prior, but local offsets caused by geometric inaccuracies persist. TSD-SR-FT+FP further incorporates the camera-aware feature projector into TSD-SR-FT, preserving the fine-detail recovery of the TSD-SR prior while producing target-view-aligned results.

Tab.~\ref{tab:ablation} further shows the complementary roles of the two components. The gains between the first two rows in the columns under Focal Length Only indicate that the TSD-SR prior primarily benefits focal-length control, whereas the gains between the last two rows in the Viewpoint Only columns show that the camera-aware feature projector primarily benefits viewpoint control. These quantitative results corroborate the qualitative observations in Fig.~\ref{fig:ablation}.

To further illustrate the camera-aware feature projector's ability to produce camera-aligned features, Fig.~\ref{fig:feature_projection_module} visualizes its output features using PCA, with the first three principal components mapped to RGB. The two rows, corresponding to different target views, show that the projected features adapt to each target view according to its viewpoint and focal length, providing robust feature-space guidance for the generative backbone.

\section{Conclusion}
In this paper, we introduced single-image rephotography, the task of synthesizing new shots of the same scene from a single reference image with specified viewpoints, focal lengths, and parameterized photographic effects. We presented ReX-Shot as a unified solution to this task. To mitigate projection bias caused by imperfect single-image geometry, we combine pixel-space explicit projection with feature-space implicit projection to provide complementary target-view guidance. To recover details lost under focal-length enlargement, we formulate focal-length control as a geometry-guided super-resolution problem and leverage the detail prior of a one-step generative super-resolution model. Building on the resulting 3D-aware generative backbone, we further introduce dedicated photographic-effect encoders for continuous, content-consistent effect control across viewpoints and focal lengths. Experiments demonstrate that ReX-Shot outperforms representative baselines across viewpoint, focal-length, and photographic-effect control tasks. Beyond these individual control tasks, ReX-Shot supports joint control within a unified generation process and enables near-real-time interaction through its one-step generative backbone.

Despite these strengths, the current scope of ReX-Shot is limited to focal-length enlargement, a relatively narrow range of viewpoint changes, and one photographic effect per forward pass. Future work will broaden this scope by supporting shorter focal lengths, larger viewpoint changes, and compositional control of multiple photographic effects.


%





\ifCLASSOPTIONcaptionsoff
  \newpage
\fi



%



\bibliographystyle{IEEEtran}
\bibliography{egbib}

%

\begin{IEEEbiography}[{\includegraphics[width=1in,height=1.25in,clip,keepaspectratio]{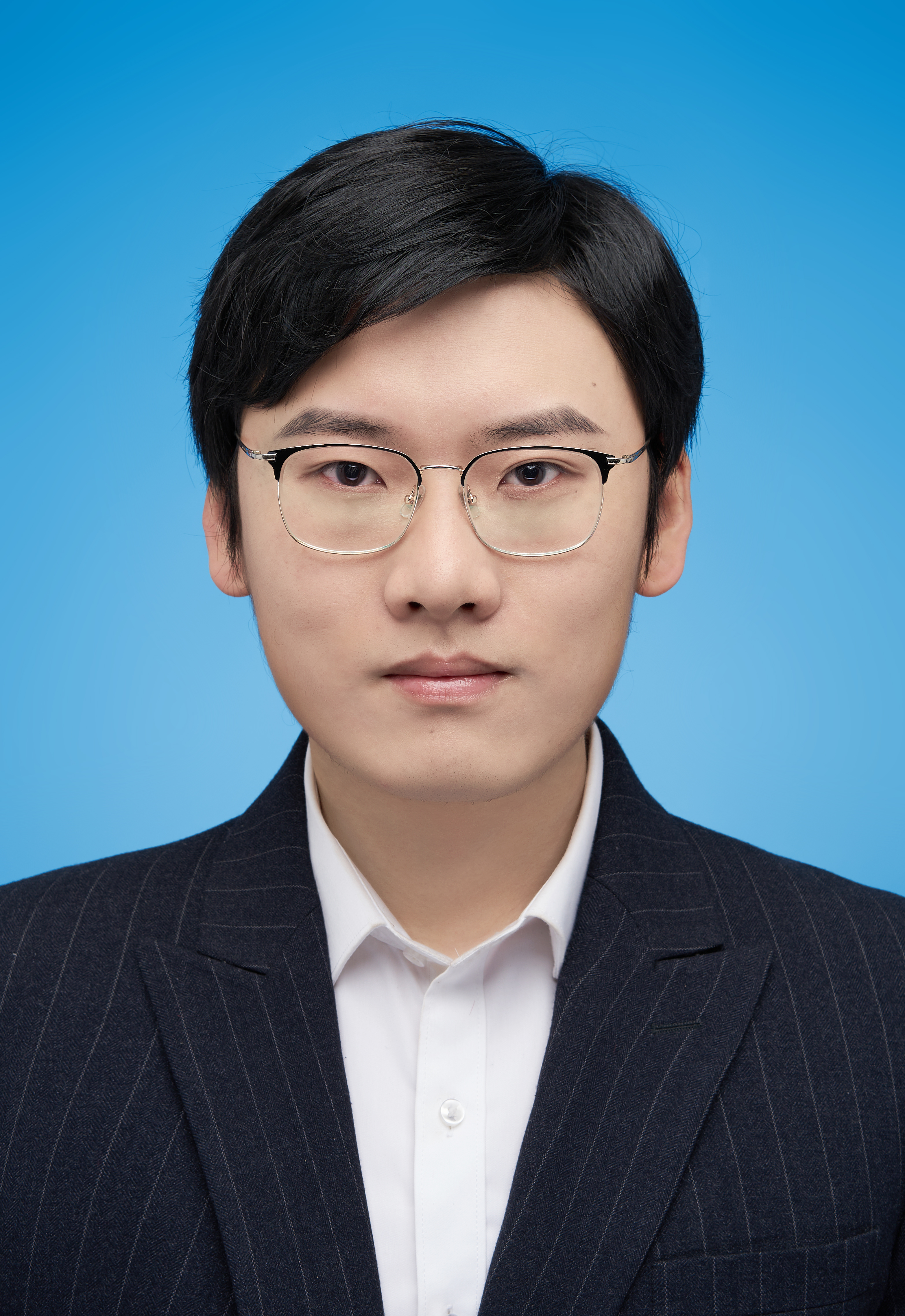}}]{Ruiqi Zhang} received his B.S. degree from the School of Electronic and Optical Engineering, Nanjing University of Science and Technology, Nanjing, China, in 2022. He is currently pursuing the Ph.D. degree with the School of Electronic Science and Engineering, Nanjing University, Nanjing, China. His current research interests include computational photography and 3D reconstruction and perception.
\end{IEEEbiography}

\begin{IEEEbiography}[{\includegraphics[width=1in,height=1.25in,clip,keepaspectratio]{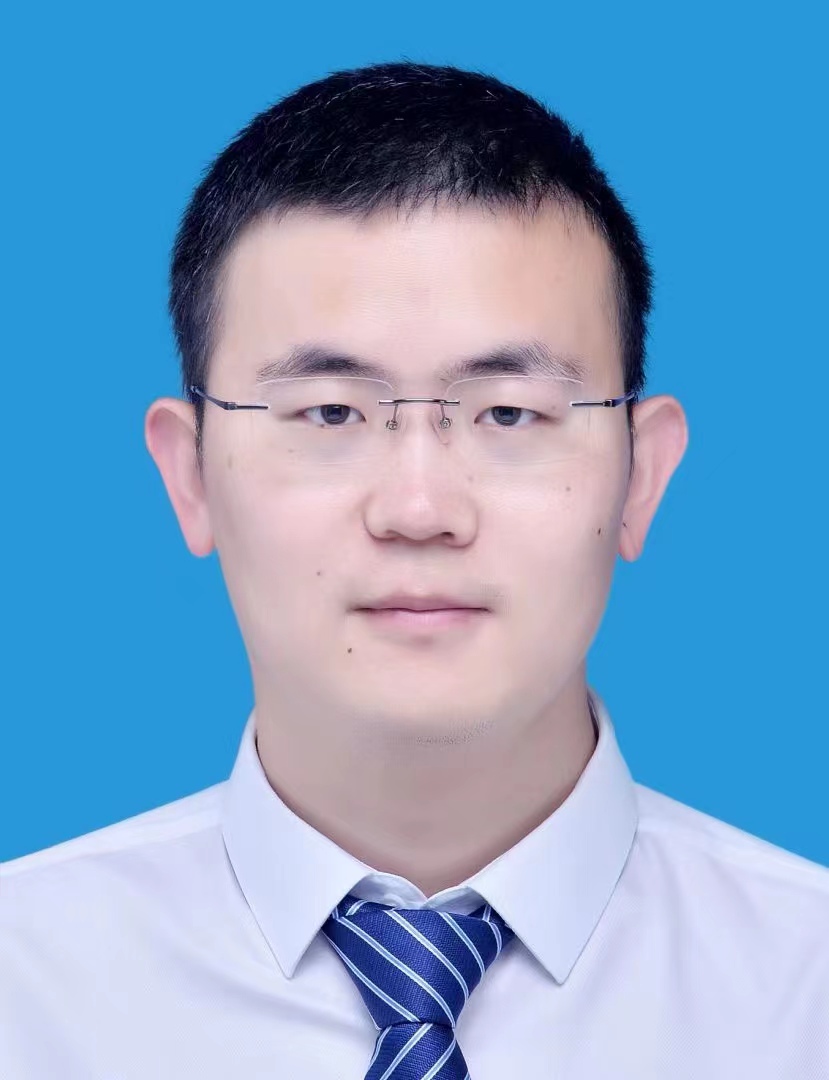}}]{Hao Zhu} is an Assistant Professor in the School of Electronic Science and Engineering, Nanjing University. He received the B.S. and Ph.D. degrees from Northwestern Polytechnical University in 2014 and 2020, respectively. He was a visiting scholar at the Australian National University. His research interests include computational photography and optimization for inverse problems.
\end{IEEEbiography}

\begin{IEEEbiography}[{\includegraphics[width=1in,height=1.25in,clip,keepaspectratio]{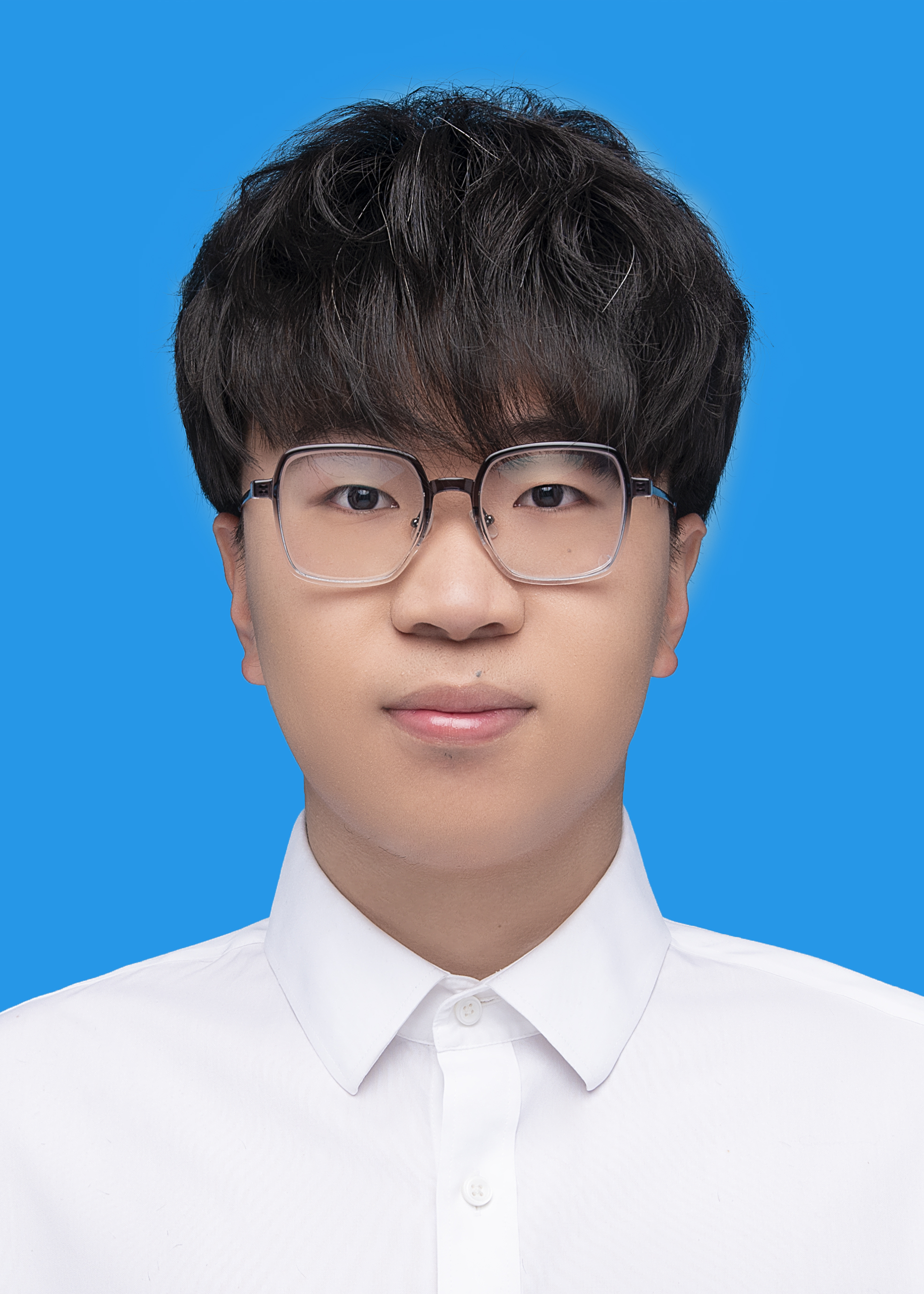}}]{Wenhao Zhang} received a B.S. degree from Nankai University, Tianjin, China. He is currently pursuing the M.S. degree with the School of Electronic Science and Engineering, Nanjing University, Nanjing, China. His research interests include 3D reconstruction and generation.
\end{IEEEbiography}


\begin{IEEEbiography}[{\includegraphics[width=1in,height=1.25in,clip,keepaspectratio]{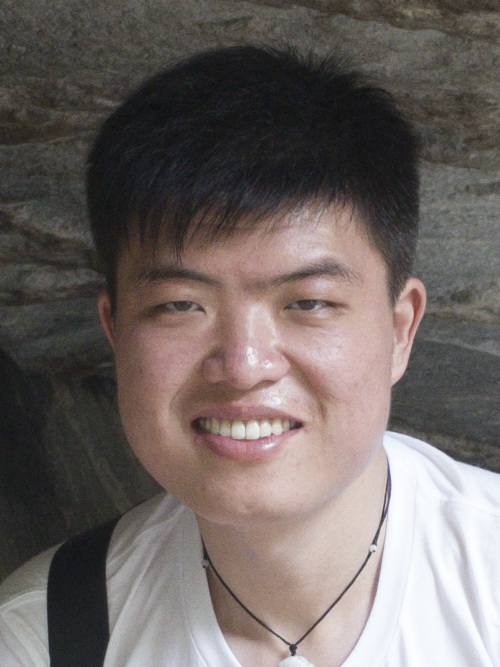}}]{Qi Zhang} is currently a lead researcher with vivo Mobile Communication Co., Ltd. in Xi'an, China. Before that, he was a researcher with Tencent AI Lab. He received his Ph.D. from the School of Computer Science at Northwestern Polytechnical University in 2021. He received CCF Doctorial Dissertation Award Nominee in 2021. His research interests include 3D vision, neural rendering, Gaussian Splatting, and AIGC.
\end{IEEEbiography}

\begin{IEEEbiography}[{\includegraphics[width=1in,height=1.25in,clip,keepaspectratio]{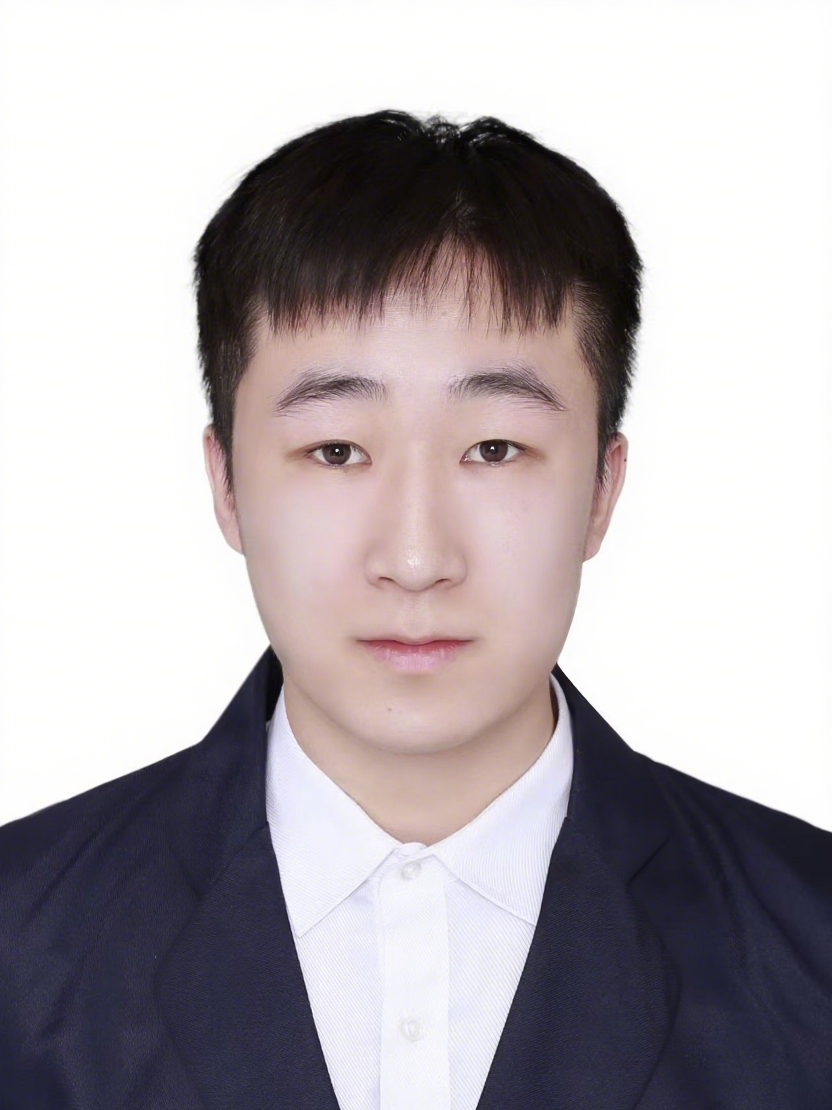}}]{Junqi Shi} received his B.S. and M.E. degrees in the School of Electronic Science and Engineering from Nanjing University, Nanjing, China in 2022 and 2025 respectively, where he is currently pursuing the Ph.D. degree. His current research interests include deep learning-based image/video coding, model quantization and signal representation.
\end{IEEEbiography}

\begin{IEEEbiography}[{\includegraphics[width=1in,height=1.25in,clip,keepaspectratio]{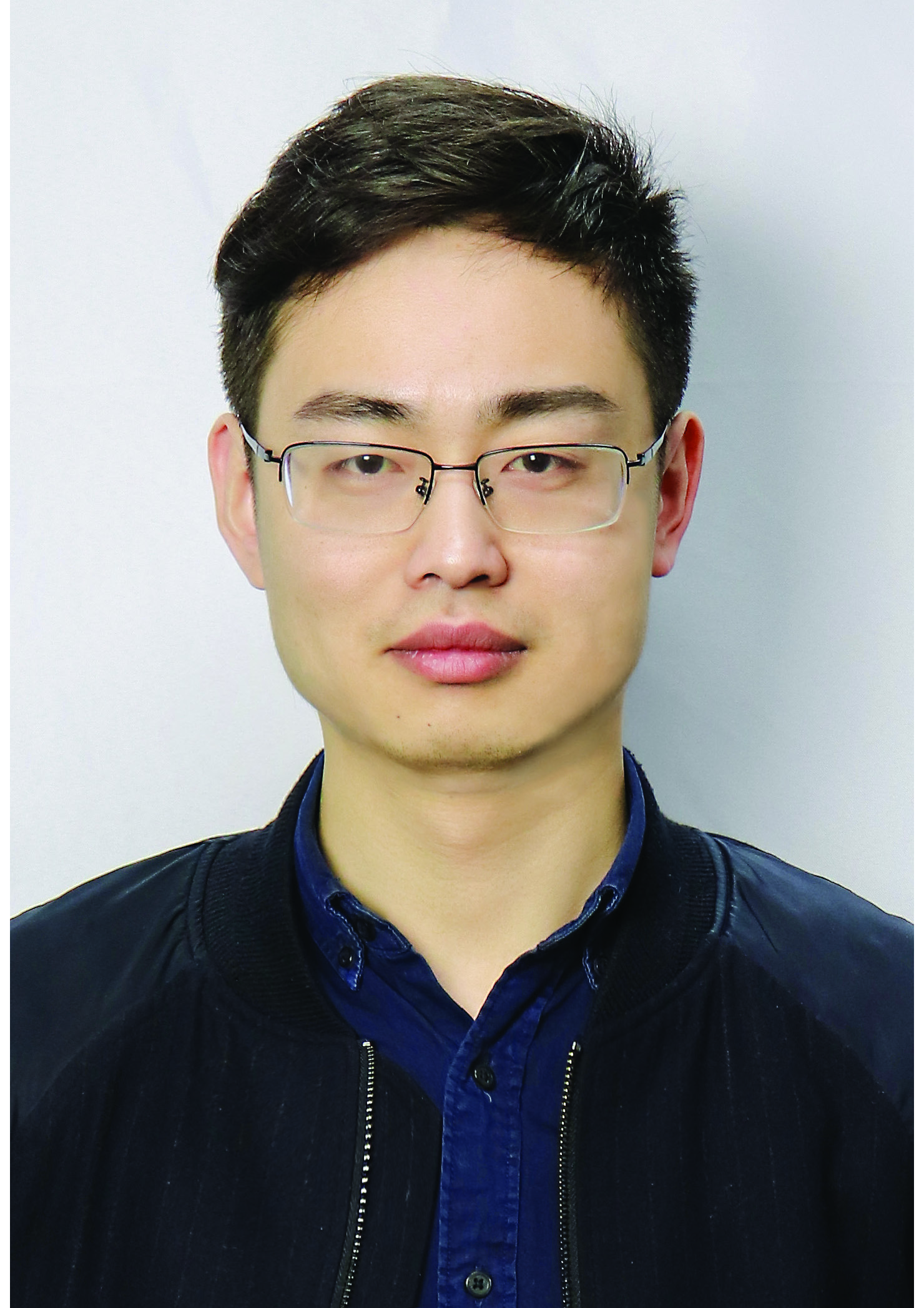}}]{Ming Lu} is an Associate Researcher with the School of Electronic Science and Engineering, Nanjing University, Nanjing, China. He received his B.S. and Ph.D. degrees in the School of Electronic Science and Engineering from Nanjing University, Nanjing, China, in 2016 and 2023 respectively. His research focuses on deep learning-based image/video coding. He is a co-recipient of the 2018 ACM SIGCOMM Student Research Competition Finalist, the 2020 IEEE MMSP Image Compression Grand Challenge Best Performing Solution, and the 2023 IEEE WACV Best Algorithms Paper Award.
\end{IEEEbiography}

\begin{IEEEbiography}[{\includegraphics[width=1in,height=1.25in,clip,keepaspectratio]{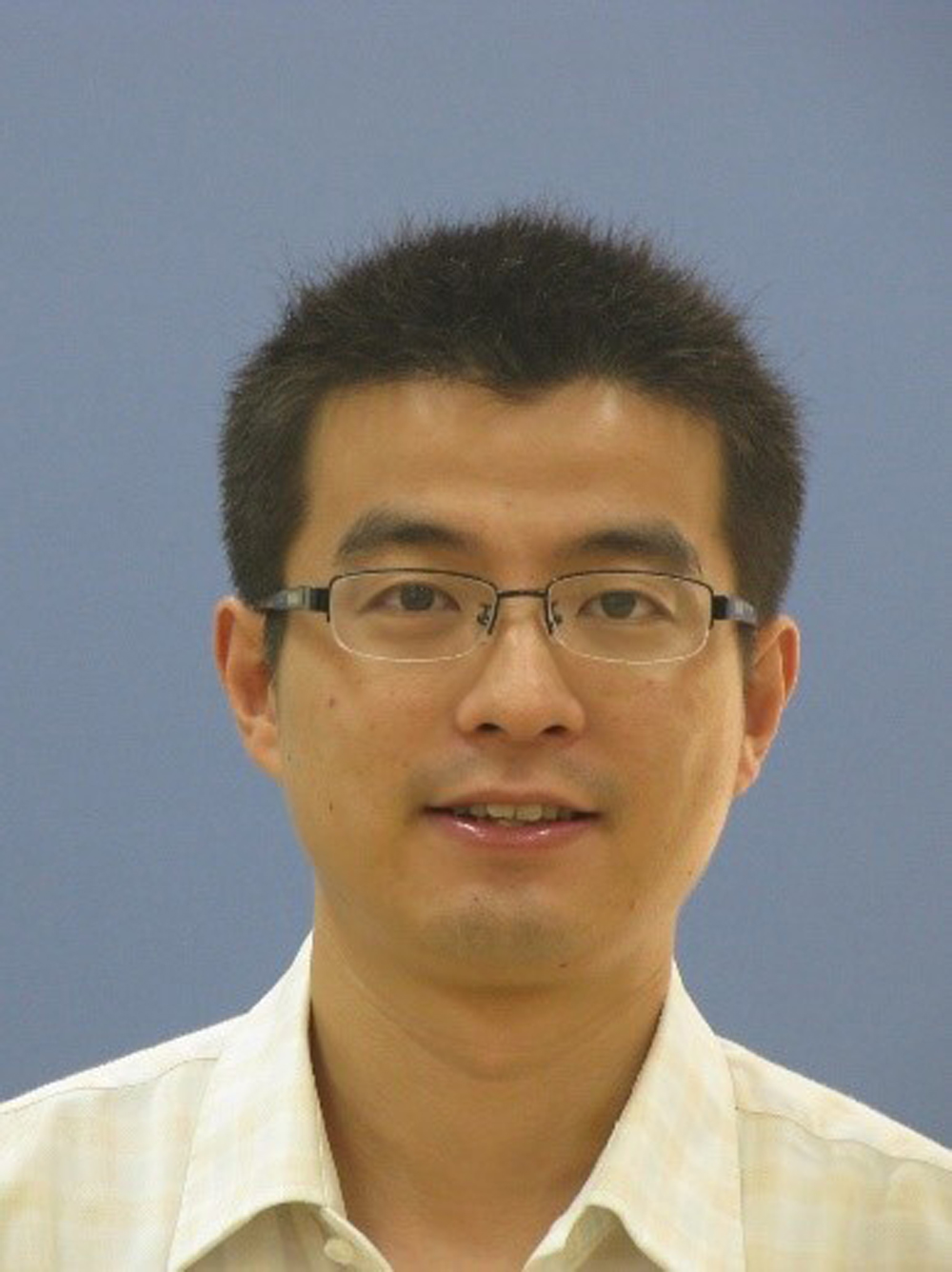}}]{Xun Cao} received the B.S. degree from Nanjing University, Nanjing, China, in 2006, and the Ph.D. degree from the Department of Automation, Tsinghua University, Beijing, China, in 2012. He held visiting positions with Philips Research, Aachen, Germany, in 2008 and Microsoft Research Asia, Beijing, from 2009 to 2010. He was a Visiting Scholar with the University of Texas at Austin, Austin, TX, USA, from 2010 to 2011. He is a Professor at the School of Electronic Science and Engineering, Nanjing University. His current research interests include computational photography and image-based modeling and rendering.
\end{IEEEbiography}

\begin{IEEEbiography}[{\includegraphics[width=1in,height=1.25in,clip,keepaspectratio]{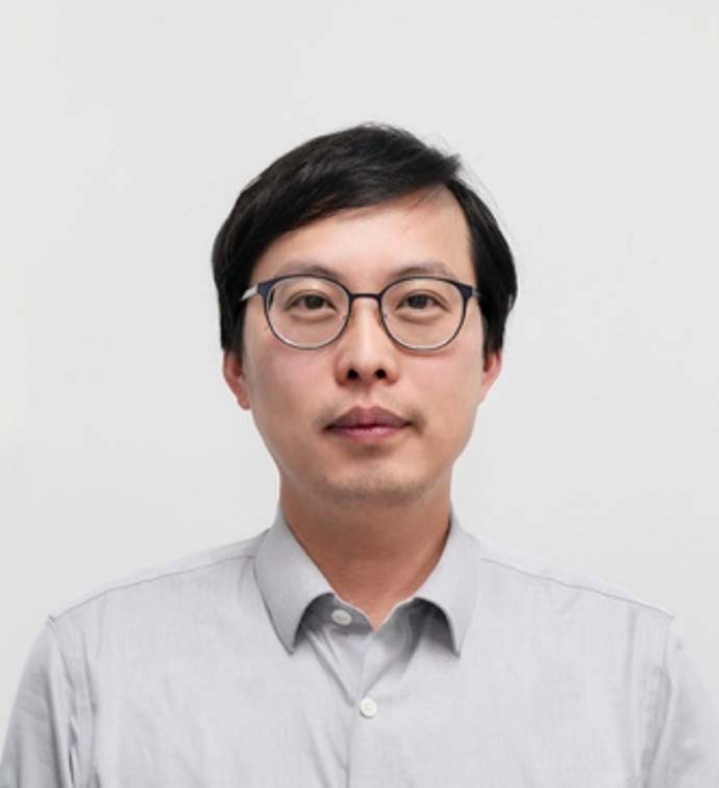}}]{Zhan Ma} (SM'19) is a Professor in Electronic Science and Engineering School, Nanjing University, Nanjing, Jiangsu, 210093, China. He received the B.S. and M.S. degrees from the Huazhong University of Science and Technology, Wuhan, China, in 2004 and 2006, respectively, and the Ph.D. degree from the New York University, New York, in 2011. From 2011 to 2014, he has been with Samsung Research America, Dallas, TX, and  Futurewei Technologies, Inc., Santa Clara, CA, respectively. His research focuses on learning-based video communication and computational imaging. He is a co-recipient of the 2019 IEEE Broadcast Technology Society Best Paper Award, the 2020 IEEE MMSP Image Compression Grand Challenge Best Performing Solution, the 2023 IEEE WACV Best Algorithm Paper Award, and the 2023 IEEE Circuits and Systems Society Outstanding Young Author Award.
\end{IEEEbiography}




\end{document}